\documentclass[lettersize,journal]{IEEEtran}
\usepackage{amsmath,amsfonts}
\usepackage{algorithmic}
\usepackage{algorithm}
\usepackage{array}
\usepackage[caption=false,font=normalsize,labelfont=sf,textfont=sf]{subfig}
\usepackage{textcomp}
\usepackage{stfloats}
\usepackage{url}
\usepackage{verbatim}
\usepackage{graphicx}
\usepackage{cite}
\usepackage{bm}
\usepackage{multirow}
\usepackage{booktabs}
\usepackage{xcolor} 
\usepackage{colortbl}
\usepackage{makecell}
\usepackage{pifont}
\usepackage{subfig}
\usepackage{amssymb}

\begin{document}



\title{Uni-MDTrack: Learning Decoupled Memory and Dynamic States for Parameter-Efficient Visual Tracking in All Modality}

\author{Wenrui Cai, Zhenyi Lu, Yuzhe Li, Yongchao Feng, Jinqing Zhang, \\ Qingjie Liu,~\IEEEmembership{Member, IEEE}, Yunhong Wang,~\IEEEmembership{Fellow, IEEE}
\thanks{All authors are with the School of Computer Science and Engineering, Beihang University, Beijing 100191, China. Corresponding to Qingjie Liu (qingjie.liu@buaa.edu.cn) and Yunhong Wang (yhwang@buaa.edu.cn)}}

\markboth{Journal of \LaTeX\ Class Files,~Vol.~14, No.~8, August~2021}%
{Shell \MakeLowercase{\textit{et al.}}: A Sample Article Using IEEEtran.cls for IEEE Journals}


\maketitle

\begin{abstract}


With the advent of Transformer-based one-stream trackers that already possess strong capability in inter-frame relation modeling, recent research has increasingly focused on how to introduce spatio-temporal context into trackers.
However, most existing methods only rely on a limited number of historical frames, which not only leads to insufficient utilization of spatio-temporal context, but also inevitably increases the length of input sequences and incurs prohibitive computational overhead. Methods that query an external memory bank, on the other hand, often suffer from inadequate fusion between the retrieved spatio-temporal features and the backbone features. Moreover, using discrete historical frames as  context overlooks the rich dynamics arising from the continuous evolution of target states in videos. 
To address the above issues, we propose a simple, powerful, and efficient spatio-temporal visual tracking framework, Uni-MDTrack, which consists of two core components: Memory-Aware Compression Prompt (MCP) module and Dynamic State Fusion (DSF) module. MCP effectively compresses rich memory features into memory-aware prompt tokens, which deeply interact with the input sequence throughout the entire backbone, significantly enhancing model performance while maintaining a stable computational load. DSF complements the discrete memory features by capturing the continuous dynamic state of the target, progressively introducing the updated dynamic state features from shallow to deep layers of the tracker, while also preserving high operational efficiency. 
Beyond the superior capacity for spatio-temporal context learning, Uni-MDTrack also supports unified tracking across RGB, RGB-D/T/E, and RGB-Language modalities. More importantly, experiments show that in Uni-MDTrack, training only the MCP, DSF, and prediction head, keeping the proportion of trainable parameters below 30\%, yields substantial performance gains, achieves state-of-the-art results on 10 datasets spanning five modalities. Furthermore, both MCP and DSF exhibit excellent generality, functioning as plug-and-play components that can boost the performance of various baseline trackers, while significantly outperforming existing parameter-efficient training approaches.
\end{abstract}


\section{Introduction}
\IEEEPARstart{M}{odern} SOT methods\cite{ye_2022_joint,Zhu_2023_CVPR_vipt,Wu_2023_CVPR_dropmae,LoRAT,fu2022sparsett} adopt one-stream paradigm, leveraging Transformer-based backbones \cite{dosovitskiy2020image} to process both the template and search region simultaneously using self-attention.
Despite the strong capacity for template-search region feature extraction and relation modeling, 
vanilla one-stream trackers have not yet effectively incorporated video context during tracking, and meanwhile suffer from prohibitive training overhead that requires extensive training epochs on large-scale datasets. Consequently, a growing number of works have investigated how to introduce spatiotemporal context, especially under the constraint of using only a small number of trainable parameters and training epochs. 

\begin{figure}[t]
\centering
\includegraphics[scale=0.5]{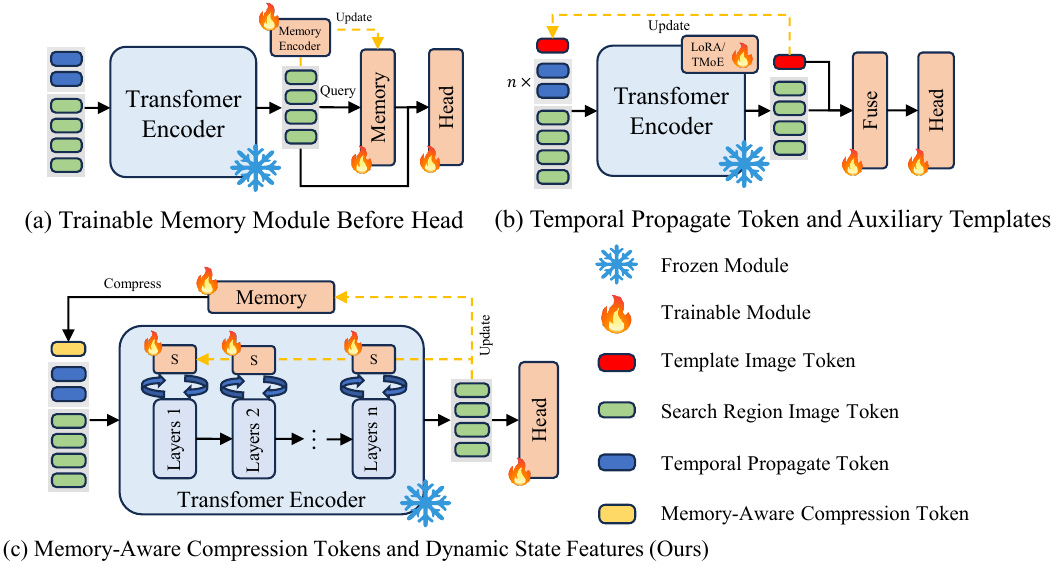}
\caption{
Comparison of existing methods for introducing spatio-temporal context features to one-stream trackers, including the parameter-efficient fine-tuning paradigms.
(a) Introducing a memory bank, where the memory features are fused before the prediction head.
(b) Introducing auxiliary templates and temporally propagated tokens.
(c) Introducing memory-aware compressed tokens as prompts at the model input, and incorporating dynamic state features of the target into the multi-stage layers of the backbone.
}
\label{fig:1}
\end{figure}

Existing methods in this direction can be broadly categorized into two classes, as shown in Figure \ref{fig:1}. Methods corresponding to Figure \ref{fig:1}(a) introduces a memory bank to store more historical features~\cite{Cai_2024_CVPR_HIPTrack,fu2021stmtrack}, and performs adaptive querying based on the features of the current search region. 
However, these methods update the memory at fixed frame intervals, and cannot effectively handle drastic, short-term target variations. Meanwhile, memory features are not introduced until the prediction head, which results in a lack of deep fusion with the search region features.
As for methods corresponding to the category shown in Figure \ref{fig:1}(b)~\cite{yan2021learning,Cui_2022_MixFormer,he2023target_TATrack,Xie_2024_CVPR_AQATrack}, they introduce multiple auxiliary historical frames into the input sequence of the backbone, enabling full fusion of historical features with current search region features within the backbone. Additionally, some other methods~\cite{ODTrack,Cai_2025_CVPR_SPMTrack} leverage temporally propagated tokens to capture the dynamic features of the tracking target. However, these components have inherent limitations. The propagated token interacts with both the template and search region tokens simultaneously, which leads to a significant portion of its attention being focused on the template area, making it function more as a template enhancer rather than a representation of continuous state changes of the target. 
Furthermore, the sparsely sampled auxiliary templates neglect a vast amount of contextual information and risk introducing distractors, while substantially increasing computational costs due to the extended input length. 
At present, both categories include representative parameter-efficient training methods. For example, HIPTrack~\cite{Cai_2024_CVPR_HIPTrack} falls into the first category and requires training only the memory encoder and the prediction head, whereas SPMTrack~\cite{Cai_2025_CVPR_SPMTrack} belongs to the second category and trains only the LoRA or TMoE modules in the backbone along with the prediction head. However, the inherent limitations of these two categories in spatio-temporal context modeling have not been resolved.
The limitations motivate our question: \textit{How to introduce rich memory and continuous dynamic state features of the target to trackers via a parameter-efficient training paradigm, while ensuring both the overall operational efficiency, and the deep fusion between spatio-temporal and search region features?}

To address the question, in this paper, we propose Uni-MDTrack, a simple, powerful and efficient spatio-temporal visual tracking framework, which consists of two core components: Memory-Aware Compression Prompt module (MCP) and Dynamic State Fusion module (DSF). 
As shown in Figure \ref{fig:1}(c), similar to memory-based methods, MCP also maintains a memory bank. The key distinction lies in that MCP employs dynamic queries to compress memory bank into a fixed set of memory-aware tokens. By concatenating these tokens to the input sequence, MCP achieves deep interaction between memory, template and search region features with a minimal increase in sequence length, thereby preserving computational efficiency. Meanwhile, memory-aware tokens also alleviate the problem of limited contextual information contained in auxiliary templates.

For the DSF module, we adopt State Space Models (SSMs)~\cite{ssm} to continuously update the dynamic state of the target. DSF only leverages search region features for state updating, which ensures full capture of the dynamic changes of the target while avoiding the issue that temporal propagated tokens may be contaminated by excessive template information.  
Moreover, we apply DSF at multiple stages of the backbone, thereby ensuring deep fusion between the dynamic features and the backbone representations.
Notably, DSF is fundamentally different from previous tracking approaches that apply SSMs~\cite{lai2025mambavt,li2025mambalct,Liu_2025_CVPR_mambavlt}. Previous methods utilize SSM as an implementation of the trainable backbone, utilizing the linear computational complexity of SSM to extend context length. However, these approaches require designing complex
scanning algorithms, and SSMs are also not guaranteed to outperform Transformers under the same sequence length~\cite{10.5555/3692070.3693514_illusion_of_SSM}. In contrast, our DSF continuously utilizes new search region features to update target states, and employs these states as supplementary features for the backbone, eliminating the need to design scanning strategy. The key focus of the DSF module is to demonstrate that continuously updated dynamic target states can serve as effective features, rather than being constrained to specific model designs like SSM. 

Furthermore, rather than re-architecting or replacing backbone layers, we inject memory and dynamic
states with MCP and DSF in a lightweight adapter manner,
eliminating the need of full-parameter training. 
As shown in Figure \ref{fig:1}(c), in Uni-MDTrack, only the MCP and DSF modules, together with the prediction head, are trained, while all other modules can remain frozen.
Meanwhile, to fully exploit the strong spatio-temporal modeling capability of Uni-MDTrack, we integrate tracking across all modalities (RGB, RGB-T, RGB-E, RGB-D, RGB-Language) into a single Uni-MDTrack model, supporting a more comprehensive set of modalities than previous multimodal tracking methods~\cite{Zhu_2023_CVPR_vipt,Hou_2024_CVPR_sdstrack,hu2025exploiting_sttrack,Wu_2024_CVPR_untrack}. Experimental results demonstrate that training under 30\% of its parameters for only 50 epochs, Uni-MDTrack achieves state-of-the-art performance on 10 datasets, including LaSOT~\cite{fan2019lasot}, TrackingNet~\cite{muller2018trackingnet}, VisEvent~\cite{10284004_visevent}, and Depthtrack~\cite{Yan_2021_ICCV_depthtrack}. Moreover, our proposed MCP and DSF modules demonstrate excellent generalization ability, acting as plug-and-play enhancements that effectively boost the performance of diverse trackers.




To summarize, our contributions are as follows: \textbf{(1)} 
We propose Memory-Aware Compression Prompt module (MCP)
and Dynamic State Fusion module (DSF) to efficiently and effectively introduce rich memory features and continuous dynamic state features.
\textbf{(2)}  
Based on MCP and DSF, We propose Uni-MDTrack, a simple, powerful and efficient spatio-temporal
visual tracking framework. Uni-MDTrack support tracking across all modalities, and only contains under 30\% trainable parameters, while
ensuring both the overall operational efficiency, and the deep
fusion between spatio-temporal and search region features.
\textbf{(3)} 
Experimental results demonstrate that Uni-MDTrack achieves state-of-the-art performance on 10 datasets across all modalities. 
Furthermore, our proposed MCP and DSF modules can be used as plug-and-play components to effectively enhance the performance of various other models.

\section{Related Work}

\subsection{Trackers with Spatio-Temporal Context}

Methods such as STARK~\cite{yan2021learning}, MixFormer~\cite{Cui_2022_MixFormer}, and TrDiMP~\cite{wang2021transformer}  samples several historical frames as auxiliary templates. However, sparsely sampled frames overlook rich context and are prone to introduce distractors. Auxiliary templates also significantly increase the input sequence length, leading to an increase in computational overhead. 
Previous methods such as HIPTrack~\cite{Cai_2024_CVPR_HIPTrack}, to ensure computational efficiency, only fuse memory features with search region features after the backbone network, and additionally design a dedicated memory feature encoder. This results in more complex network architectures and insufficient fusion between memory features and search region features.
Other methods like ODTrack~\cite{ODTrack}, SPMTrack~\cite{Cai_2025_CVPR_SPMTrack} and AQATrack~\cite{Xie_2024_CVPR_AQATrack} introduce temporal propagated tokens across consecutive frames. MambaLT~\cite{li2025mambalct}, TemTrack~\cite{xie2025robust_temtrack}, and STTrack~\cite{hu2025exploiting_sttrack} further collect temporal propagation tokens from multiple search regions and collectively enhance them using SSMs like Mamba~\cite{mamba}. However, the limitation is that the propagation tokens attend to both the template and the search region, leading to the template diverting the attention of propagation tokens away from the search region and hindering the effective expression of target dynamic states. The enhancement applied to these tokens also primarily serves to strengthen template features as well. Other methods like MambaVT~\cite{lai2025mambavt} and MCITrack~\cite{kang2025exploring_mcitrack} utilize Mamba as an implementation of the trainable backbone. However, as analyzed in Introduction, unlike previous SSM methods, our proposed DSF continuously utilizes new frame search region features to update target states and employs these states as supplementary information for the backbone. 
The key focus of the DSF module is to demonstrate that continuously updated dynamic target states can serve as effective features, rather than being constrained to specific model designs such as SSM.
To the best of our knowledge, we are also the first to leverage SSM as a PEFT technique in visual tracking.
%

\begin{figure*}
    \centering
    \includegraphics[width=\textwidth]{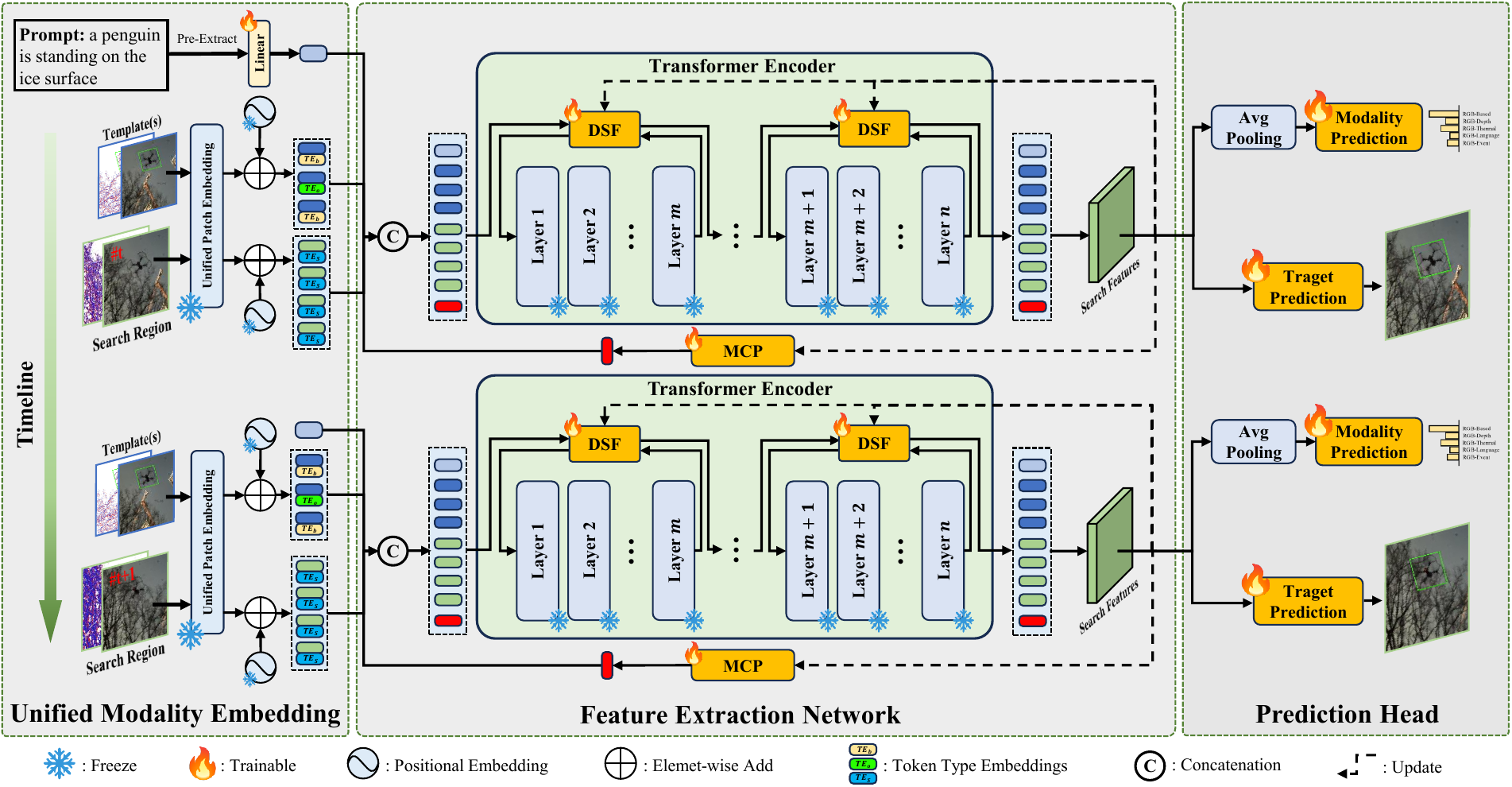}
    \vspace{-2ex}
    \caption{The overall architecture of Uni-MDTrack. Uni-MDTrack can uniformly process data from various modalities and consists of unified modality embedding layer, feature extraction network and prediction head. }
    \label{fig:2}
\end{figure*}

\subsection{Parameter-Efficient Fine-tuned Trackers}

Most current trackers adopt one-stream paradigm~\cite{ye_2022_joint, Cui_2022_MixFormer,Wu_2023_CVPR_dropmae,chen2022backbone}.
However, 
one-stream trackers demand significantly more training steps and employ large backbones such as ViT-L~\cite{dosovitskiy2020image}, leading to a substantial training costs. Thanks to abundant RGB data and the class-agnostic nature of the SOT task, trackers possess strong generalization capabilities. Consequently, a growing body of research has shifted towards PEFT to enhance the performance of existing trackers. 
PEFT methods can be broadly divided into two categories. The first category introduces auxiliary modalities to supplement an RGB-based foundation tracker (\textit{e.g.} RGB-D, RGB-T, RGB-E). ProTrack~\cite{yang2022prompting_protrack}, ViPT~\cite{Zhu_2023_CVPR_vipt}, and SeqTrackV2~\cite{chen2023unified_seqtrackv2} incorporate lightweight modules to extract features from the auxiliary modality while keeping the backbone of the foundation tracker frozen. SDSTrack~\cite{Hou_2024_CVPR_sdstrack} and OneTracker~\cite{Hong_2024_CVPR_onetracker} simultaneously fine-tunes foundation tracker and a fusion module to extract modality-enhanced features. However, these approaches still demonstrate weaker performance than unified multi-modal foundation trackers trained from scratch, such as SUTrack~\cite{sutrack} and FlexTrack~\cite{tan2025you_flextrack}.
The second category of fine-tuning methods focus on strengthening the capabilities of the tracker itself, particularly spatio-temporal context modeling abilities. HIPTrack~\cite{Cai_2024_CVPR_HIPTrack} introduces historical features by training historical prompt network. LoRAT~\cite{LoRAT} maintains a strong generalization by fine-tuning DINOv2~\cite{oquab2024dinov2}. SPMTrack~\cite{Cai_2025_CVPR_SPMTrack}  further incorporates TMoE, auxiliary templates, and temporal propagation tokens for fine-tuning. 
In contrast, our proposed Uni-MDTrack not only supports tracking across all modalities, but also introduces more comprehensive and powerful spatio-temporal context features in a parameter-efficient manner.

\section{Method}

\subsection{Overall Architecture}

As illustrated in Figure~\ref{fig:2}, we present \textbf{Uni-MDTrack}, a novel visual tracking framework built upon a prompt module based on Memory-Aware Compression Prompt module (MCP) and Dynamic State Fusion module (DSF), which supports tracking across all modalities: pure RGB, RGB-D, RGB-E, RGB-T, and RGB-Language.
The overall architecture follows a one-stream paradigm, primarily consisting of a unified modality embedding layer, a feature extraction network based on HiViT~\cite{zhang2023hivit}, and two prediction heads for target prediction and modality prediction.
Throughout the model, the backbone remains frozen, with only MCP, DSF, and the prediction head serving as the main modules involved in training.

Specifically, input images from different modalities are first processed through the unified patch embedding module to generate a unified representation embedding. Positional embeddings and token type embeddings 
are then added to the unified representation embeddings. For text encoding in RGB-Language tracking tasks, we pre-extract the [\emph{cls}] token as the text embedding using the pretrained text encoder from CLIP-L \cite{pmlr-v139-radford21a}.
All embedded tokens are fed into the feature extraction network. 
Within the feature extraction network, the memory-aware compression tokens output by MCP are first concatenated with the input tokens and input into the backbone. The backbone processes all tokens simultaneously, while performing fusion with dynamic state features via DSF from shallow to deep layers.
Finally, we employ a center-based prediction head ~\cite{ye_2022_joint} to predict the tracking result 
 , while a task recognition head~\cite{sutrack} is used to predict the modality of the current input, thereby better assisting the model in capturing task-specific features.

\subsection{Unified Modality Embedding}

We adopt the same unified patch embedding module as \cite{sutrack}. The unified patch embedding layer modifies the conventional patch embedding by extending the input dimension of linear projection layer from 3 to 6. 
RGB images are denoted as 
\(
\bm{X}_{\text{RGB}} \in \mathbb{R}^{H \times W \times 3},
\)
while depth, thermal, and event images are collectively referred to as DTE. We replicate DTE images into 3-channel images and normalize each pixel value to the range of [0, 255] to obtain
\(
\bm{X}_{\text{DTE}} \in \mathbb{R}^{H \times W \times 3}.
\)
We then construct a 6-channel input by concatenating the RGB and DTE image along the channel dimension to obtain $\bm{X} \in \mathbb{R}^{H\times W \times 6}$. For RGB and RGB-language tasks that do not include DTE images, we duplicate the RGB channels to form the required 6-channel input $\bm{X}$. Through the unified patch embedding process, template and search region images are transformed into respective token sequences $\bm{T} \in \mathbb{R}^{N_T \times d}$ and $\bm{S} \in \mathbb{R}^{N_S \times d}$. All tokens $\bm{T}$ and $\bm{S}$ are then added with positional embeddings soft token type embeddings following \cite{sutrack}. For the RGB-Language tracking, the text feature token is projected through a small linear layer to obtain $\bm{L} \in \mathbb{R}^{1\times d}$.
All tokens are concatenated and fed into the feature extraction network.

\subsection{Feature Extraction Network}

Our feature extraction framework is by default built upon HiViT~\cite{zhang2023hivit}. 
The feature extraction network first concatenates the input tokens with the memory-aware compression tokens $\bm{M} \in \mathbb{R}^{N_M \times d}$ output by the MCP module, resulting in the input $\bm{Z} \in \mathbb{R}^{N\times d}$ to the backbone. 
The backbone processes the sequence and continuously integrates dynamic state features from DSF modules. 
The final output sequence $\bm{O} \in \mathbb{R}^{N \times d}$ of the feature extraction network is used for the final target prediction. Meanwhile, output tokens corresponding to search region, denoted as $\bm{O_S}$, are added to the memory bank in MCP, as well as update the state in DSF modules.

\subsubsection{Memory-Aware Compression Prompt Module (MCP)} \

The design of MCP is guided by two principles. First, introduce memory features at the input of the backbone, thereby enabling deep interaction with the template and search region features. Second, effectively compress memory features, thus maintaining a stable computational load for the overall model. 
Based on the above principles, we propose a dynamic query-based resampling approach for memory feature compression. Specifically, as shown in Figure \ref{fig:3}(a), given the features $\bm{F}_m \in \mathbb{R}^{N_{mb} \times d}$ stored in the memory bank, MCP contains a total of $N_M$ trainable query tokens $\bm{q} \in \mathbb{R}^{N_M\times d}$, which perform dynamic querying and adaptive aggregation on $\bm{F}_m$ via these query tokens. The process can be formally described as follows:

\begin{figure}[t]
    \centering
    \includegraphics[width=\linewidth]{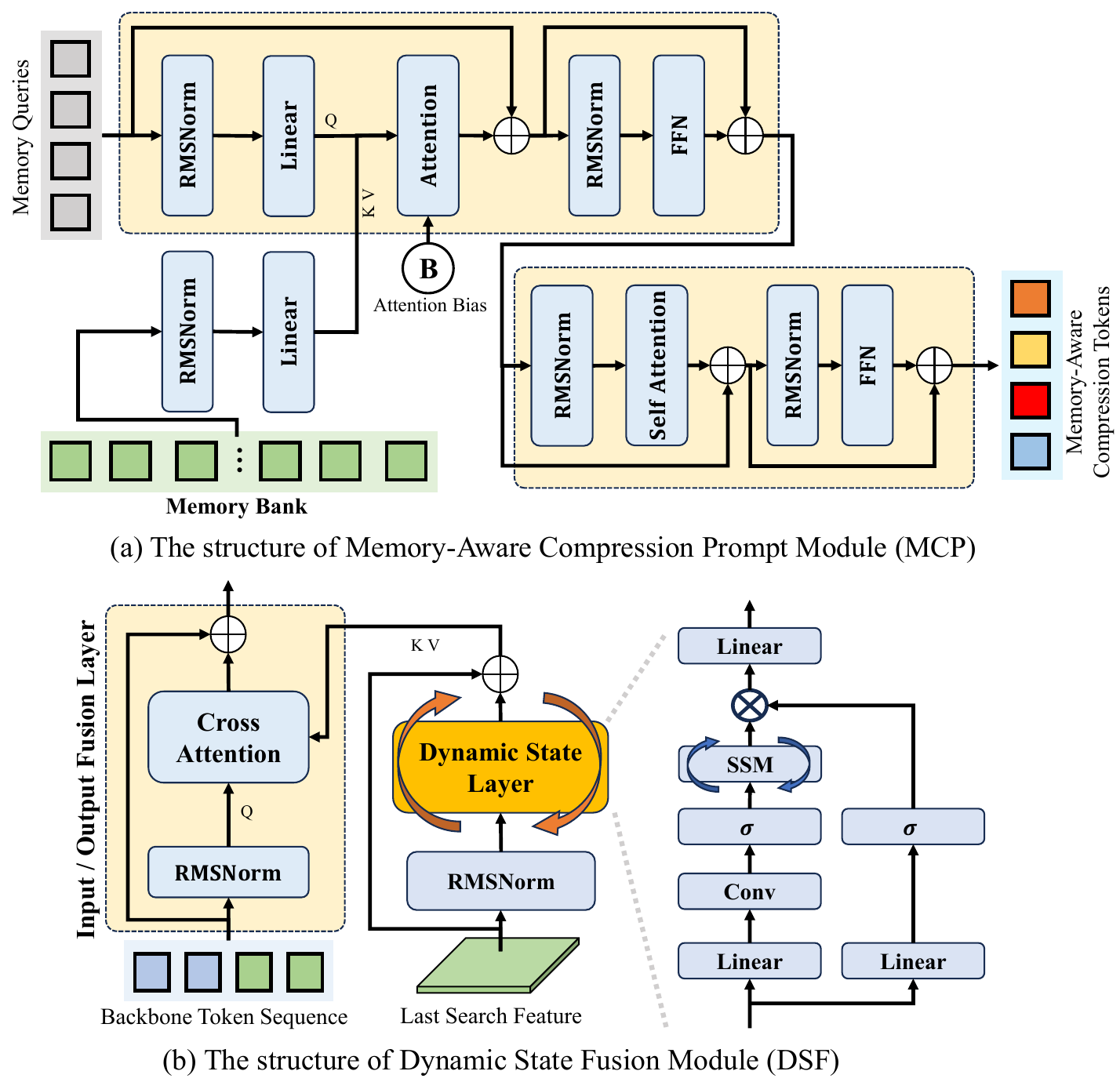}
    \vspace{-4ex}
    \caption{Detail structure of our proposed Memory-Aware Compression Prompt module (MCP) and Dynamic State Fusion module (DSF).}
    \label{fig:3}
    \vspace{-2ex}
\end{figure}

\begin{equation}
\small
\begin{aligned}
\bm{Q} =  \mathrm{Linear}_q(\mathrm{RMS}&\mathrm{Norm}_q(\bm{q})) \\
\bm{K}, \bm{V} = \mathrm{Split}(\mathrm{Linear}_{kv}&(\mathrm{RMSNorm}_{kv}(\bm{F}_m)))  \\
\bm{Attn} = \mathrm{Softmax}[\frac{\bm{Q} \cdot \bm{K}}{\sqrt{d}}& + \mathrm{ALiBi}(\bm{F}_m)] \\
\bm{M}_1 = \mathrm{Linear}_o(\bm{Attn}& \cdot \bm{V}) + \bm{q}  \\
\bm{M}_2 = \mathrm{FFN}(\mathrm{RMSNorm}&(\bm{M}_1)) + \bm{M}_1
\end{aligned}
\label{eq1}
\end{equation}

\noindent where $\bm{Attn}\in\mathbb{R}^{N_M\times N_{mb}}$ is the attention weight, $\bm{M}_2 \in \mathbb{R}^{N_M \times d}$ is the memory features after query aggregation. For clarity, Equation \ref{eq1} presents a simplified single-head attention formulation, although our actual implementation employs a multi-head attention mechanism. We also introduce an attention bias term $\mathrm{ALiBi}(\bm{F}_m) \in \mathbb{R}^{N_{mb}}$~\cite{presstrain_alibi}, which provides $\bm{F}_m$ with extrapolatable positional information. ALiBi not only allows for significantly larger memory during inference but also effectively prioritizing recent memories over older, visually similar ones and mitigating the lingering effects of potential distractors.
We encodes ALiBi position at the frame level; assuming a token $\bm{F}_m^i$ originates from the $j^{th}$ frame in the memory bank, its corresponding bias is $-\bm{m}_h \times |j-N_{mb}|$, where $h$ is the index of attention head, and each head is associated with a unique slope $\bm{m}_h=2^{\frac{-8}{h}}$. 
To prevent unbounded memory growth during inference, when the number of tracked frames exceeds $\bm{L}$, we uniformly sample the search region tokens of $\bm{L}$ frames from the tracked frames  to serve as the memory bank.
After obtaining the memory feature $\bm{M}_2$, we further use a self-attention module and a FFN layer to enhance it and output the final memory-aware compression tokens $\bm{M}$.


\subsubsection{Dynamic State Fusion Module (DSF)} \


The DSF module is conceived to meet two fundamental requirements: possessing enough capacity for capturing continuous target dynamic states, and enabling deep integration with the backbone. Therefore, we introduce the DSF module at multiple stages of the backbone from shallow to deep layers, enabling sufficient and deep fusion between the target dynamic state features and the backbone representations. By default, DSF is implemented with an SSM. Compared with the temporal propagated tokens used in prior methods, the larger state space of the SSM provides a stronger capacity to represent the dynamic states. But DSF is not limited to being implemented with an SSM.

As shown in Figure \ref{fig:3}(b), DSF module consists of three key components: an input fusion layer, a dynamic state layer based on SSM, and an output fusion layer. 
The input fusion and output fusion layer respectively integrate the target dynamic states feature $\bm{F}$ output by DSF with the input $\bm{Z^i}$ of the $i^{th}$ backbone layer and output $\bm{O}^j$ of the $j^{th}$ layer.
We can flexibly configure the number of DSF modules and choose which network layers to fuse with the DSF output features.
The DSF module utilizes only the search region features $\bm{O_S} \in \mathbb{R}^{N_S\times d}$ from the output of feature extraction network to perform state update, excluding the influence of the template or other tokens, thereby allowing the model to specifically capture the dynamics of the target itself.
As shown in Figure \ref{fig:3}(b), the overall process of the dynamic state layer can be formally described as:
\begin{equation}
\small
\begin{aligned}
\bm{I} &=  \mathrm{RMSNorm}(\bm{O_S}) \\
\bm{G} &= \mathrm{SiLU}(\mathrm{Linear}_g(\bm{I})) \\
\bm{S_{1}} &= \mathrm{SiLU}[\mathrm{Conv}(\mathrm{Linear}_c(\bm{I}))] \\
\bm{S} &= \mathrm{SSM}(\bm{S_1}) \\
\bm{F} &= \bm{O_S} + \mathrm{Linear}(\bm{G} \odot \bm{S}) 
\end{aligned}
\end{equation}

\noindent The overall process follows a gated structure, where $\bm{G} \in \mathbb{R}^{N_S\times d_s}$ represents the gating values, and $d_s$ is the inner dimension of the dynamic state layer. $\bm{S_{1}} \in \mathbb{R}^{N_S \times d_s}$ is the input to the SSM after a linear projection and convolution-based activation, and $\bm{S} \in \mathbb{R}^{N_S \times d_s}$ is the output of the SSM. An element-wise multiplication of $\bm{G}$ with $\bm{S}$, followed by a linear layer, restores the dimensionality to $d$, yielding $\bm{F} \in \mathbb{R}^{N_S \times d}$.
The state update and output of the SSM can be formally described as:
\begin{equation}
\small
\begin{aligned}
\bm{\Delta}', &\bm{B}, \bm{C} = \mathrm{Split}(\mathrm{Linear}_1(\bm{S}_1)) \\
\bm{\Delta} &= \text{Softplus}(\text{Linear}_2(\bm{\Delta}')) \\
\bar{\bm{A}} &= \exp(\bm{\Delta} \bm{A}), \bar{\bm{B}} = \bm{\Delta} \times \bm{B} \\
h(t) &= \bar{\bm{A}} \odot h(t-1) + \bar{\bm{B}} \odot \bm{S}_1, \\
\bm{S} &= \bm{C} h(t) + \bm{D} \odot \bm{S}_1
\end{aligned}
\label{eq_ssm}
\end{equation}

\noindent where $\bm{A}=-\text{exp}(\bm{A}_\text{log}) \in \mathbb{R}^{d_s \times e}$ and $\bm{D} \in \mathbb{R}^{d_s}$ are learnable parameters, $e$ is the state dimension, $\bm{\Delta} \in \mathbb{R}^{N_S \times d_s}$, $\bm{B},\bm{C}\in \mathbb{R}^{N_S \times e}$. 
$\times$ denotes the outer product operation, $\bm{\bar{B}} \in \mathbb{R}^{N_S \times d_s \times e}$, and the entire process adheres to the discrete-time formulation of SSM. $h(t) \in \mathbb{R}^{N_S\times d_s \times e}$ denotes the hidden state at time $t$, with $h(0)$ initialized as a zero matrix. $\odot$ denotes the channel-wise multiplication. The SSM uses the current feature $\bm{S_1}$ to update $h(t)$ and generate the dynamic state $\bm{S}$ of the target.
After obtaining $\bm{F}$ from the dynamic state layer, as shown in Figure \ref{fig:3}(b), the input fusion layer is designed based on cross attention,  integrates $\bm{F}$ with $\bm{Z^i}$ to produce the new input $\bm{Z^{i'}}$ of the $i^{th}$ backbone layer. The output fusion layer maintains an identical structure to the input fusion layer, integrating $\bm{F}$ with $\bm{O^j}$ to produce the new output $\bm{O^{j'}}$ of the $j^{th}$ backbone layer.

\subsection{Prediction Head}

The prediction head encompasses two tasks: target prediction and input modality classification. For target prediction, we first extract search region features $\bm{O_S}$ from the output $\bm{O}$ of the feature extraction network and feed $\bm{O_S}$ into the center-based prediction head \cite{ye_2022_joint}. For input modality classification, we apply global average pooling to the output $\bm{O_S}$ and apply an MLP for classification.

\subsection{More Theoretical Analysis on MCP and DSF}

\subsubsection{Analysis of The Limitations of SSMs in Long-Sequence Extrapolation} \ 

Formally, the hidden state update in SSM is defined as:
$h(t) = \bar{\bm{A}} \odot h(t-1) + \bar{\bm{B}} \odot \bm{S}_1$, where  $\bar{\bm{A}}$ represents the channel-wise decay factors, and each element in  $\bar{\bm{A}} \in (0,1)$. 
Because $\bm{\Delta} = \mathrm{Softplus}(\text{Linear}_2(\bm{\Delta}')) > 0, \bm{A}<0$.
Consequently, every dimension of $\bar{\bm{A}}$ is strictly less than 1. By unrolling the recurrence, the contribution of an early token $X_j$ to the output at position $i$ is proportional to the cumulative product of decay factors:
$\alpha_{i,j} \propto \left(\prod_{k=j+1}^{i} \bar{\bm{A}} \right)\odot \bar{\bm{B}}.$
Leveraging $\bar{\bm{A}} = \exp(\bm{\Delta}\bm{A})$, this product collapses into a unified exponential term:
$\prod_{k=j+1}^{i}\bar{\bm{A}} = \exp\left(\sum_{k=j+1}^{i}\bm{\Delta}\bm{A}\right)$.
Since $\bm{A}<0$ and $\bm{\Delta}>0$, there exists a constant $c>0$ such that the magnitude of influence decays exponentially with distance:
$\left\|\prod_{k=j+1}^{i}\bar{\bm{A}}\right\| \le \exp\big(-c\,(i-j)\big).$
This derivation leads to a direct conclusion: the influence of early tokens vanishes exponentially as the distance $i-j$ increases. While this decay may be manageable within the training length $L_{\text{train}}$, it becomes catastrophic during inference when extrapolating to $L_{\text{test}} \gg L_{\text{train}}$.  Therefore, directly using an SSM as the backbone and simply extending the context length—as done in prior methods—will inevitably diminish the influence of earlier tokens. This is precisely why we employ an SSM‑like structure only within DSF to model short-term dynamic state features, rather than introducing long‑term memory.

\subsubsection{Analysis of the Impact of ALiBi and Increasing Memory Length During Inference.} \ 

Let the current query be $q_t$ and the memory bank contain keys $(k_i)$ indexed by $i$. The attention scores can be defined as:
$a_{t,i} = \frac{q_t^\top k_i}{\sqrt d}+\beta\,\Delta(i,t)$
where $\Delta(i,t)$ is the relative distance, and $\beta <0$ is ALiBi slope. 
Thus, the cross-attention logit and softmax weight are
$p_{t,i} = \frac{e^{a_{t,i}}}{\sum_{j\in\mathcal M} e^{a_{t,j}}}$, where $\mathcal{M}$ denotes memory bank. 
Considering a simplified case where feature similarities are negligible ($q_t^\top k_i \approx q_t^\top k_j=0$), the relative attention weight between two memories $i$ and $j$ depends solely on their distance:
$\frac{p_{t,i}}{p_{t,j}} = \exp\big(\beta[\Delta(i,t)-\Delta(j,t)]\big).$
If memory $i$ is more recent than $j$ (i.e., $\Delta(i,t) < \Delta(j,t)$), then $p_{t,i} > p_{t,j}$. 

Assume the model is trained with a memory length $K$, but tested with length $L > K$. The total attention mass contributed by the unseen ``tail'' (memories beyond distance $K$) is bounded by a geometric series:
$\text{Mass}_{\text{tail}} = \sum_{k=K+1}^{L} e^{\beta k} <  \sum_{k=K+1}^{\infty} e^{\beta k} = \frac{e^{\beta(K+1)}}{1-e^{\beta}}.$
To ensure the trained model's attention distribution remains valid during inference, we require this tail mass to be less than a threshold $\eta$. Solving $\frac{e^{\beta(K+1)}}{1-e^{\beta}} \le \eta$ for $K$ yields:
$K \;\gtrsim\; \frac{\ln(1/\eta)}{|\beta|} - 1.$
This result implies that the effective memory horizon is of order $O(1/|\beta|)$. 
Consequently, extending the memory bank size $L$ at test time adds only an exponentially small tail to the distribution, ensuring that our MCP module extrapolates robustly.

\section{Experiments}

\label{exp}

\subsection{Implementation Details}

\noindent\textbf{Model settings.} 
We propose two versions of our tracker, Uni-MDTrack-B and Uni-MDTrack-L. Uni-MDTrack-B employs a template size of $112\times112$ and a search region size of $224\times 224$, while Uni-MDTrack-L uses $196\times196$ and $384 \times 384$, respectively. The cropping factors for the template and search region are 2.0 and 4.0 for both versions. Uni-MDTrack-B and Uni-MDTrack-L adopt HiViT-B and HiViT-L~\cite{zhang2023hivit} as backbones, respectively, and are initialized with the same weights as SUTrack-B and SUTrack-L~\cite{sutrack}. MCP module outputs a total of 16 memory-aware compression tokens, and the number of attention heads is kept consistent with backbone. We employ a total of four DSF modules. For Uni-MDTrack-B, we evenly divide the last 24 layers of HiViT-B into four stages, and fuse the dynamic state features produced by the four DSF modules with both the input and output of each stage, respectively. For Uni-MDTrack-L, we do the same for its last 40 layers. The dynamic state features are then fused with the input and output of each of these four segments.
Table \ref{table:model_settings} details the parameter and computational overhead of our models. Compared with other methods that use PEFT to enhance model capabilities, our method has a significant advantage in computational cost while introducing a comparable number of additional training parameters.

\begin{table}[h]\small
    \centering
    \caption{ Comparison of our method with other trackers using parameter-efficient training method in terms of total parameters, trainable parameters, and computational complexity.}
    \resizebox{1.0\linewidth}{!}{
    \begin{tabular}{c|ccc}
    \Xhline{2pt}
        \multirow{2}{*}{\textbf{Method}} & \textbf{Trainable} & \multirow{2}{*}{\textbf{Params(M)}} & \multirow{2}{*}{\textbf{FLOPs(G)}}  \\
         & \textbf{Params(M)} & & \\
        \Xhline{1px}
        \textbf{Uni-MDTrack-B} & 27.1 & \textbf{88.2} & \textbf{27.9} \\
        HIPTrack \cite{Cai_2024_CVPR_HIPTrack} & 34.1 & 120.4 & 66.9 \\
        LoRAT-B$_{384}$ \cite{LoRAT} & \textbf{13.0} & 99.1 & 97.0 \\
        SPMTrack-B~\cite{Cai_2025_CVPR_SPMTrack} & 29.2 & 115.3 & - \\
        \hline
        \textbf{Uni-MDTrack-L} & 54.9 & 287.4 & 257.4 \\
    \Xhline{2pt}
    \end{tabular}
    }
    \label{table:model_settings}
    \vspace{-2px}
\end{table}

\begin{table*}[!th]\footnotesize
    \centering
    \caption{State-of-the-art comparison on RGB-T tracking dataset LasHeR, RGB-E tracking dataset VisEvent, and RGB-D tracking dataset DepthTrack. 
 The best three results are highlighted in \textbf{\textcolor{red}{red}}, \textbf{\textcolor{blue}{blue}} and \textbf{bold}, respectively.}
 \setlength{\tabcolsep}{4.3mm}
    \label{whole_comparison_dte}
    \begin{tabular}{c|c|ccccccc}
        \Xhline{2pt}
        Method & Source & \multicolumn{2}{c}{LasHeR} & \multicolumn{2}{c}{VisEvent} & \multicolumn{3}{c}{DepthTrack} \\
        \cmidrule(lr){3-4}\cmidrule(lr){5-6}\cmidrule(lr){7-9} &
         & SR(\%) & PR(\%) & AUC(\%) & $P$(\%) & F-Score(\%) & Re(\%) & PR(\%)  \\
        \Xcline{1-1}{0.4pt}
        \Xhline{1pt}

        \textbf{Uni-MDTrack-B} & \textbf{Ours} & 61.2 & 76.7 & \textbf{\textcolor{blue}{64.2}} & \textbf{81.0} & 65.9 & 66.3 & 66.2 \\
        \textbf{Uni-MDTrack-L} & \textbf{Ours} & \textbf{\textcolor{red}{62.1}} & \textbf{\textcolor{red}{77.9}} & \textbf{\textcolor{red}{65.7}} & \textbf{\textcolor{red}{81.8}} & \textbf{\textcolor{red}{67.4}} & \textbf{\textcolor{red}{67.2}} & \textbf{\textcolor{red}{67.6}} \\
        FlexTrack~\cite{tan2025you_flextrack} & ICCV25 & \textbf{\textcolor{blue}{62.0}} & \textbf{\textcolor{blue}{77.3}} & \textbf{64.1} & \textbf{\textcolor{blue}{81.4}} & \textbf{\textcolor{blue}{67.0}} & \textbf{\textcolor{blue}{66.9}} & \textbf{\textcolor{blue}{67.1}} \\
        SUTrack-B$_{224}$~\cite{sutrack} & AAAI25 & 59.9 & 74.5 & 62.7 & 79.9 & 65.1 & 65.7 & 64.5 \\
        SUTrack-L$_{384}$~\cite{sutrack} & AAAI25 & \textbf{61.9} & \textbf{76.9} & 63.8 & 80.5 & \textbf{66.4} & \textbf{66.4} & \textbf{66.5} \\
        STTrack~\cite{hu2025exploiting_sttrack} & AAAI25 &60.3 & 76.0 & 61.9 & 78.6 & 63.3 & 63.4 & 63.2 \\
        SeqTrackV2-B$_{256}$~\cite{chen2023unified_seqtrackv2} & Arxiv23 & 55.8 & 70.4 & 61.2 & 78.2 & 63.2 & 63.4 & 62.9 \\
        UnTrack~\cite{Wu_2024_CVPR_untrack} & CVPR24 & 53.6 &66.7& 58.9&75.5&61.2 & 61.0 & 61.3\\
        SDSTrack~\cite{Hou_2024_CVPR_sdstrack} & CVPR24 & 53.1 &66.5&59.7&76.7&61.4 &60.9 & 61.9\\
        OneTracker~\cite{Hong_2024_CVPR_onetracker} & CVPR24 & 53.8 & 67.2 & 60.8 & 76.7 & 60.9 & 60.4 & 60.7\\
        ViPT~\cite{Zhu_2023_CVPR_vipt} & CVPR23 & 52.5 & 65.1 & 59.2 & 75.8 & 59.4 &59.6& 59.2\\
        \Xhline{2pt}
    \end{tabular}
    \vspace{-2ex}
\end{table*}

\begin{table*}[!t]
    \centering
    \caption{State-of-the-art comparison on RGB visual tracking datasets LaSOT, TrackingNet and LaSOT$_\mathrm{ext}$. 
    The best three results are highlighted in \textbf{\textcolor{red}{red}}, \textbf{\textcolor{blue}{blue}} and \textbf{bold}, respectively.}
    \footnotesize
    \label{whole_comparison}
    \begin{tabular}{c|c|ccccccccc}
        \Xhline{2pt}
        Method & Source & \multicolumn{3}{c}{LaSOT} & \multicolumn{3}{c}{TrackingNet} & \multicolumn{3}{c}{LaSOT$_\mathrm{ext}$} \\
        \cmidrule(lr){3-5}\cmidrule(lr){6-8}\cmidrule(lr){9-11} &
         & AUC(\%) & $P_{Norm}$(\%) & $P$(\%) & AUC(\%) & $P_{Norm}$(\%) & $P$(\%) & AUC(\%) & $P_{Norm}$(\%) & $P$(\%)  \\
        \Xcline{1-1}{0.4pt}
        \Xhline{1pt}
        \multicolumn{4}{l}{\emph{Unified Trackers}} \\
        \hline
        \textbf{Uni-MDTrack-B} & \textbf{Ours} & 74.7 & 84.9 & 82.6 & 86.1 & 90.8 & 85.9  & 54.3 & \textbf{\textcolor{blue}{65.7}} & \textbf{\textcolor{blue}{62.4}} \\
        \textbf{Uni-MDTrack-L} & \textbf{Ours} & \textbf{\textcolor{red}{76.1}} & \textbf{\textcolor{red}{85.7}} & \textbf{\textcolor{red}{84.3}} & \textbf{\textcolor{red}{88.0}} & \textbf{\textcolor{red}{92.1}} & \textbf{\textcolor{red}{89.1}} & \textbf{\textcolor{red}{55.2}} & \textbf{\textcolor{red}{66.3}} & \textbf{\textcolor{red}{62.8}} \\
        SUTrack-B$_{224}$ \cite{sutrack} & AAAI25 & 73.2 & 83.4 & 80.5 & 85.7  & 90.3 & 85.1 & 53.1 & 64.2 & 60.5 \\
        SUTrack-L$_{384}$ \cite{sutrack} & AAAI25 & 75.2 & 84.9 &  83.2 & \textbf{\textcolor{blue}{87.7}} & \textbf{\textcolor{blue}{91.7}} &  \textbf{\textcolor{blue}{88.7}} & 53.6 & 64.2 & 60.5 \\
        
        \hline
        \multicolumn{4}{l}{\emph{RGB-based Trackers}} \\
        \hline
        \textbf{Uni-MDTrack-B$_{\text{RGB}}$} & \textbf{Ours} & \textbf{\textcolor{blue}{75.6}} & \textbf{85.1} & \textbf{\textcolor{blue}{83.8}} & \textbf{86.4} & 90.6 & \textbf{86.2} & \textbf{\textcolor{blue}{54.8}} & \textbf{65.6} & \textbf{62.1} \\
        SPMTrack-B~\cite{Cai_2025_CVPR_SPMTrack} & CVPR25 & 74.9 & 84.0 & 81.7 & 86.1 & 90.2 & 85.6 & - & - & -\\
        ARPTrack$_{256}$~\cite{Liang_2025_CVPR_arptrack} & CVPR25 & 72.6 & 81.4 & 78.5 & 85.5 & 90.0 & 85.3 & 52.0 & 62.9 & 58.7 \\
        MCITrack-B~\cite{kang2025exploring_mcitrack} & AAAI25 & \textbf{75.3} & \textbf{\textcolor{blue}{85.6}} & \textbf{83.3} & 86.3 & \textbf{{90.9}} & 86.1 & \textbf{54.6} & \textbf{\textcolor{blue}{65.7}} & \textbf{62.1} \\
        MambaLCT$_{384}$~\cite{li2025mambalct} & AAAI25 & 73.6 & 84.1 & 81.6 & 85.2 & 89.8 & 85.2 & 53.3 & 64.8 & 61.4 \\
        LoRAT-B$_{378}$ \cite{LoRAT} & ECCV24 & 72.9 & 81.9 & 79.1  & 84.2 & 88.4 & 83.0 & 53.1& 64.8 & 60.6 \\
        AQATrack$_{384}$ \cite{Xie_2024_CVPR_AQATrack} & CVPR24 & 72.7 &82.9 & 80.2  & 84.8 & 89.3 & 84.3 & 52.7 & 64.2 & 60.8\\
        ARTrackV2-B$_{384}$ \cite{Bai_2024_CVPR_ARTrackV2} & CVPR24 & 73.0 & 82.0 & {79.6} & 85.7 & 89.8 & 85.5 & 52.9 & 63.4 & 59.1\\
        HIPTrack \cite{Cai_2024_CVPR_HIPTrack} & CVPR24 & 72.7 & 82.9 & 79.5 & 84.5 & 89.1 & 83.8 & 53.0 & 64.3 & 60.6 \\
        ODTrack-B \cite{ODTrack} & AAAI24 & 73.2 & 83.2& 80.6  &  85.1 & 90.1 & 84.9 & 52.4 & 63.9 & 60.1 \\
        ARTrack$_{384}$  \cite{Wei_2023_CVPR_autoregressive} & CVPR23 & 72.6 & 81.7 & 79.1 & 85.1 & 89.1 & 84.8 & 51.9 & 62.0 & 58.5 \\
        SeqTrack-B$_{384}$   \cite{Chen_2023_CVPR_seqtrack} & CVPR23 & 71.5 & 81.1 & 77.8 & 83.9 & 88.8 & 83.6 & 50.5 & 61.6 & 57.5 \\
        OSTrack$_{384}$  \cite{ye_2022_joint} & ECCV22 & 71.1 & 81.1 & 77.6 & 83.9 & 88.5 & 83.2 & 50.5 & 61.3 & 57.6 \\
        \Xhline{2pt}
    \end{tabular}
    \vspace{-2ex}
\end{table*}

\noindent\textbf{Datasets.} Following SUTrack \cite{sutrack}, our method utilizes the \emph{training} sets from LaSOT \cite{fan2019lasot}, GOT-10K \cite{huang2019got}, COCO \cite{lin2014microsoft}, TrackingNet \cite{muller2018trackingnet}, VastTrack \cite{NEURIPS2024_ec17a52e_vasttrack}, TNL2K \cite{Wang_2021_CVPR_TNL2k}, DepthTrack \cite{Yan_2021_ICCV_depthtrack}, VisEvent \cite{10284004_visevent}, and LasHeR \cite{9640453_lasher} for training. 
During training, in each batch sampling step, the probability ratios used for sampling from each dataset are set to 2:2:2:2:2:2:1:1:1.
We sample 7 frames per step, with the first 2 frames serving as templates and the latter 5 as search frames. For the image dataset COCO, we replicate single images multiple times to simulate sequential data. 

\noindent\textbf{Training and Optimization.} Our method is implemented based on PyTorch 2.3.1 and trained on 4 NVIDIA A100 GPUs. We set the batch size to 64 per GPU for Uni-MDTrack-B and 16 for Uni-MDTrack-L. Both version are trained for 50 epochs, with 100,000 frame sequences sampled from all datasets in each epoch. We employ the AdamW \cite{DBLP:conf/iclr/LoshchilovH19_AdamW} optimizer with an initial learning rate of 2e-4 for both stages, which is decreased to 2e-5 after 40 epochs. The weight decay is set to 1e-4 throughout the training process.

\noindent\textbf{Loss Function.} For target prediction, consistent with OSTrack \cite{ye_2022_joint}, we employ Generalized IoU \cite{rezatofighi2019generalized} Loss and L1 Loss to supervise bounding box prediction, and Focal Loss \cite{lin2017focal} to supervise target center point prediction. Additionally, we use Cross-Entropy Loss to compute the modality prediction loss. The loss weights for the above components are set to 2.0, 5.0, 1.0, and 1.0, respectively.

\noindent\textbf{Inference.} 
Consistent with SUTrack~\cite{sutrack}, we use 2 templates input. DSF module performs continuous state updates per frame during tracking, and the memory bank of MCP contains a total of 50 frames of uniformly sampled historical search region features. 

\subsection{Comparisons with the State-of-the-Art Methods}

To better compare with mainstream RGB trackers, we additionally trained a pure RGB tracker Uni-MDTrack-B$_{\text{RGB}}$, which is implemented based on SPMTrack-B~\cite{Cai_2025_CVPR_SPMTrack}. The prediction head of Uni-MDTrack-B$_{\text{RGB}}$ no longer introduces modality prediction. Similarly, we divided all 12 layers of the model into 4 equal stages to fuse DSF modules, and trained it only on LaSOT, GOT-10K, TrackingNet, and COCO with identical training configurations.

\noindent\textbf{LaSOT} \cite{fan2019lasot} is an \emph{RGB-based} tracking dataset constructed for long-term tracking. As shown in Table \ref{whole_comparison}, our approach achieves significant improvement compared to SUTrack-B$_{224}$ (\textbf{+1.5 AUC}) and SUTrack-L$_{384}$ (\textbf{+0.9 AUC}). Uni-MDTrack-B$_{\text{RGB}}$ also achieves significant improvement compared to SPMTrack-B (\textbf{+0.7 AUC}) and outperforms all RGB Trackers based on ViT-B~\cite{dosovitskiy2020image}.

\noindent\textbf{LaSOT$_\mathrm{ext}$} \cite{fan2019lasot} is an \emph{RGB-based} tracking dataset that has no overlaps with LaSOT \cite{fan2021lasot}. 
As shown in Table \ref{whole_comparison}, our method significantly outperforms SUTrack, and Uni-MDTrack-B$_{\text{RGB}}$ achieves the best performance among RGB trackers.

\begin{table}[h]\small
    \centering
    \caption{The performance of our method and other state-of-the-art trackers on RGB-Language Tracking dataset TNL2K. The best three results are highlighted in \textbf{\textcolor{red}{red}}, \textbf{\textcolor{blue}{blue}} and \textbf{bold}.}
    \begin{tabular}{c|ccc}
    \Xhline{2pt}
        \textbf{Method} & AUC(\%) & $P_{Norm}$(\%) & $P$(\%)  \\
        \Xhline{1pt}
        \textbf{Uni-MDTrack-B} & \textbf{{67.6}} & \textbf{\textcolor{blue}{85.2}} & \textbf{\textcolor{blue}{73.2}} \\
        \textbf{Uni-MDTrack-L} & \textbf{\color{red}{70.4}} & \textbf{\color{red}{87.4}} & \textbf{\color{red}{77.4}} \\
        SUTrack-B$_{224}$ \cite{sutrack} & 65.0 & - & 67.9 \\
        SUTrack-L$_{384}$~\cite{sutrack} & \textbf{\color{blue}{67.9}} & - & \textbf{72.1} \\
        MCITrack-B~\cite{kang2025exploring_mcitrack} & 62.9 & - & - \\
        LoRAT-B$_{378}$ \cite{LoRAT} & 59.9 & - & 63.7 \\
        ODTrack-L \cite{ODTrack} & 61.7 & - & - \\
        ARTrackV2-L$_{384}$ \cite{Bai_2024_CVPR_ARTrackV2} & 61.6 & - & - \\
        CiteTracker \cite{Li_2023_ICCV_citetracker} & 57.7 & - & 59.6 \\
        VLT \cite{NEURIPS2022_1c8c87c3_VLT} & 53.1 & - & 53.3 \\
        
    \Xhline{2pt}
    \end{tabular}
    \label{comparison_tnl2k}
\end{table}

\noindent\textbf{TrackingNet} \cite{muller2018trackingnet} is an \emph{RGB-based} large-scale tracking dataset. As shown in Table \ref{whole_comparison}, our method outperforms  SUTrack \cite{sutrack}, and Uni-MDTrack-B$_{\text{RGB}}$ outperforms other RGB trackers.

 \noindent \textbf{UAV123, OTB2015 and NfS} ~\cite{mueller2016benchmark,otb2015,Galoogahi_2017_ICCV_nfs} are \emph{RGB-based} datasets. We conduct evaluations
 on the 30 FPS version of NfS. As shown in Table \ref{table:uav,nfs,otb}, our method significantly outperforms SUTrack-B$_{384}$ with a larger resolution.

\noindent\textbf{GOT-10k}~\cite{huang2019got} is an \emph{RGB-based} dataset. Previous trackers, when tested on GOT-10k, are typically trained only on the GOT-10k dataset itself~\cite{Cai_2025_CVPR_SPMTrack,Bai_2024_CVPR_ARTrackV2,Wei_2023_CVPR_autoregressive,ye_2022_joint,fu2022sparsett}. Since our proposed Uni-MDTrack is a tracker that supports all modalities and is trained on datasets from multiple modalities, a direct comparison with previous methods on GOT-10k would not be fair. Therefore, we have provided the results of our method in Table \ref{comparison_got10k} as a reference. 

\begin{table}[!h]\small
    \centering
    \caption{The performance of our method and other state-of-the-art trackers on RGB-based Tracking dataset GOT-10k. The best three results are highlighted in \textbf{\textcolor{red}{red}}, \textbf{\textcolor{blue}{blue}} and \textbf{bold}.}
    \begin{tabular}{c|ccc}
    \Xhline{2pt}
        \textbf{Method} & AO(\%) & $SR_{0.5}$(\%) & $SR_{0.75}$(\%)  \\
        \Xhline{1pt}
        \textbf{Uni-MDTrack-B} & \textbf{\textcolor{red}{81.1}} & \textbf{\textcolor{red}{91.8}} & \textbf{\textcolor{red}{81.2}}  \\
        SUTrack-B$_{224}$~\cite{sutrack} & \textbf{\textcolor{blue}{77.9}} & \textbf{87.5} & \textbf{\textcolor{blue}{78.5}}\\
        SPMTrack-B~\cite{Cai_2025_CVPR_SPMTrack} &76.5 &85.9& 76.3 \\
        MCITrack-B~\cite{kang2025exploring_mcitrack}&\textbf{\textcolor{blue}{77.9}} &\textbf{\textcolor{blue}{88.2}}& \textbf{76.8} \\
        ARPTrack$_{256}$~\cite{Liang_2025_CVPR_arptrack} & 77.7 & 87.3 & 74.3\\
        MambaLCT$_{384}$~\cite{li2025mambalct} & 76.2 & 86.7 & 74.3 \\
        LoRAT-B$_{378}$~\cite{LoRAT} & 73.7 & 82.6 & 72.9 \\
        
    \Xhline{2pt}
    \end{tabular}
    \label{comparison_got10k}
\end{table}

 \noindent\textbf{TNL2K} \cite{Wang_2021_CVPR_TNL2k} is an \emph{RGB-Language} tracking dataset. Each video is accompanied by natural language description. As shown in Table \ref{comparison_tnl2k}, our method also significantly outperforms SUTrack-B$_{224}$  
 (\textbf{+2.6 AUC}) and SUTrack-L$_{384}$ (\textbf{+2.5 AUC}). 
 Our approach outperforming existing state-of-the-art methods by a significant gap.

\begin{table}[t]\small
    \centering
    \caption{The performance of our method and other state-of-the-art trackers on UAV123, NfS and OTB2015 in terms of AUC metrics. The best three results are highlighted in \textbf{\textcolor{red}{red}}, \textbf{\textcolor{blue}{blue}} and \textbf{bold}.}
    \setlength{\tabcolsep}{3.4mm}
    \begin{tabular}{c|ccc}
    \Xhline{2pt}
        \textbf{Method} & \textbf{UAV123} & \textbf{NfS} & \textbf{OTB2015}\\
        \Xhline{1pt}
        \textbf{Uni-MDTrack-B} & \textbf{\textcolor{red}{71.0}} & \textbf{\textcolor{red}{70.2}} & \textbf{\textcolor{red}{73.6}}  \\
        SUTrack-B$_{384}$ \cite{sutrack} & \textbf{70.4} & \textbf{\textcolor{blue}{69.3}} & - \\
        HIPTrack \cite{Cai_2024_CVPR_HIPTrack} & \textbf{\textcolor{blue}{70.5}} & \textbf{68.1} & 71.0 \\
        ARTrackV2-B \cite{Bai_2024_CVPR_ARTrackV2} & 69.9 & 67.6 & - \\
        ODTrack-L \cite{ODTrack} & - & - & \textbf{\textcolor{blue}{72.4}} \\
        ARTrack$_{384}$  \cite{Wei_2023_CVPR_autoregressive} & \textbf{\textcolor{blue}{70.5}} & 66.8 & - \\
        SeqTrack-B$_{384}$  \cite{Chen_2023_CVPR_seqtrack} & 68.6 & 66.7 & - \\
        MixFormer-L  \cite{Cui_2022_MixFormer} & 69.5 & - & - \\
        
    \Xhline{2pt}
    \end{tabular}
    \label{table:uav,nfs,otb}
\end{table}

 \noindent\textbf{DepthTrack} \cite{Yan_2021_ICCV_depthtrack} is an \emph{RGB-Depth} tracking dataset. As shown in Table \ref{whole_comparison_dte}, our method achieves state-of-the-art performance on DepthTrack.

 \noindent\textbf{LasHeR} \cite{9640453_lasher} is an \emph{RGB-Thermal} short-term tracking dataset with high-diversity. As shown in Table \ref{whole_comparison_dte}, our method achieves current best performance and shows a remarkable boost in terms of PR.

 \noindent\textbf{VisEvent} \cite{10284004_visevent} is an \emph{RGB-Event} tracking dataset. As shown in Table \ref{whole_comparison_dte}, our method achieves state-of-the-art performance, demonstrating significantly higher performance than SUTrack~\cite{sutrack}. Uni-MDTrack-B can achieve even better results than SUTrack-L$_{384}$.

\subsection{More Detailed Results in Different Attribute
Scenes on LaSOT}

\begin{figure*}[!t]
\centering
{
\begin{minipage}{4.1cm}
\centering
\includegraphics[scale=0.23]{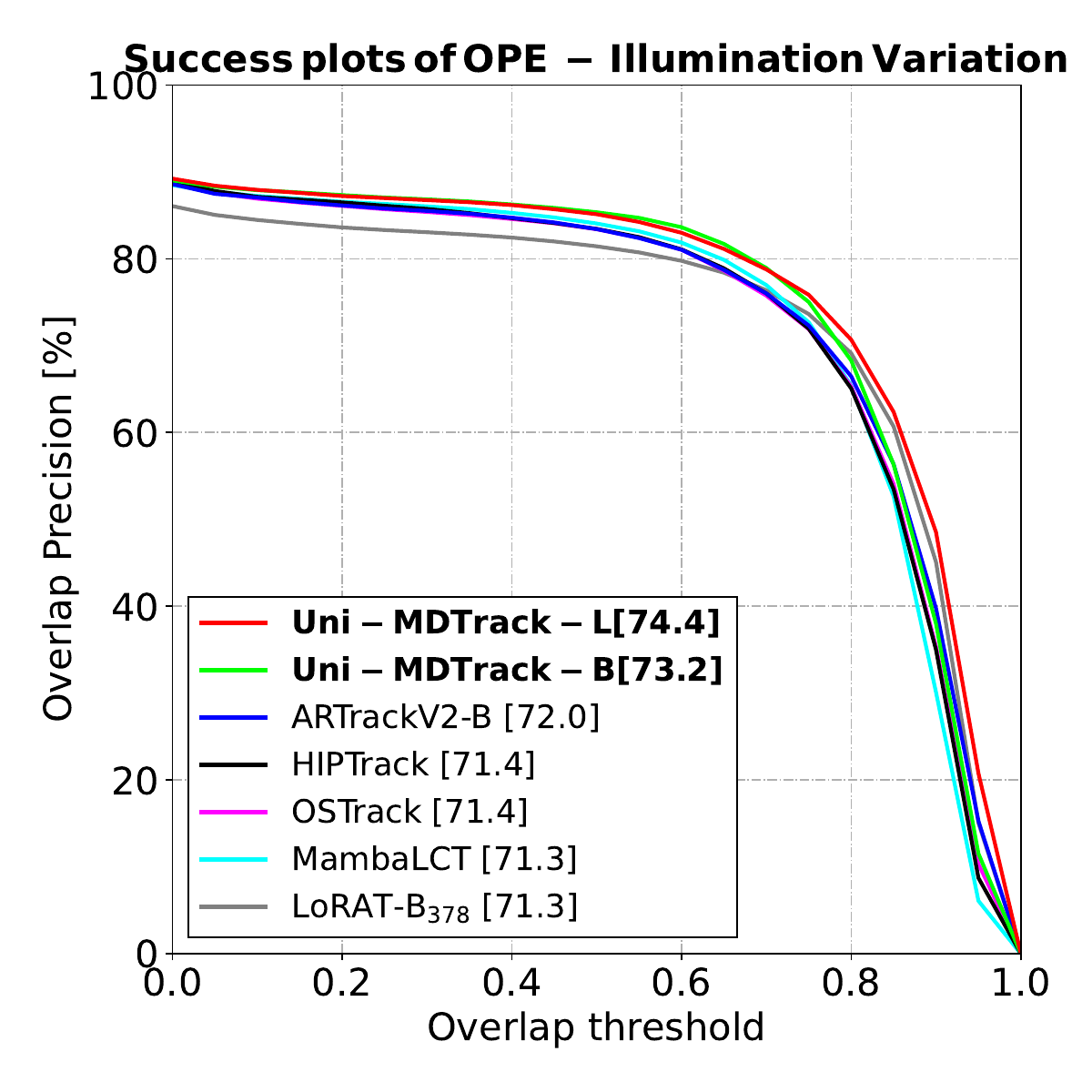} 
\end{minipage}
}
{
\begin{minipage}{4.1cm}
\centering
\includegraphics[scale=0.23]{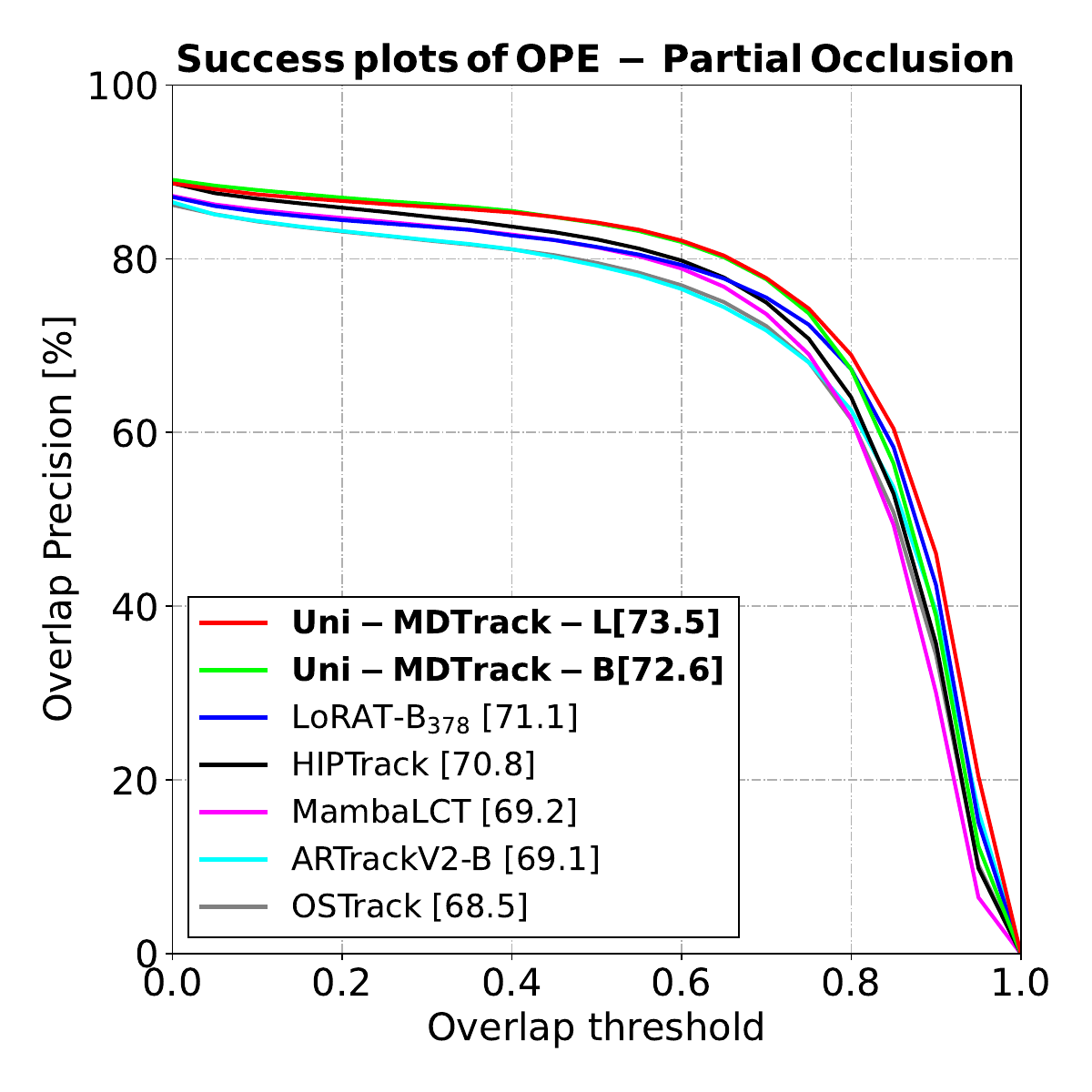}
\end{minipage}
}
{
\begin{minipage}{4.1cm}
\centering
\includegraphics[scale=0.23]{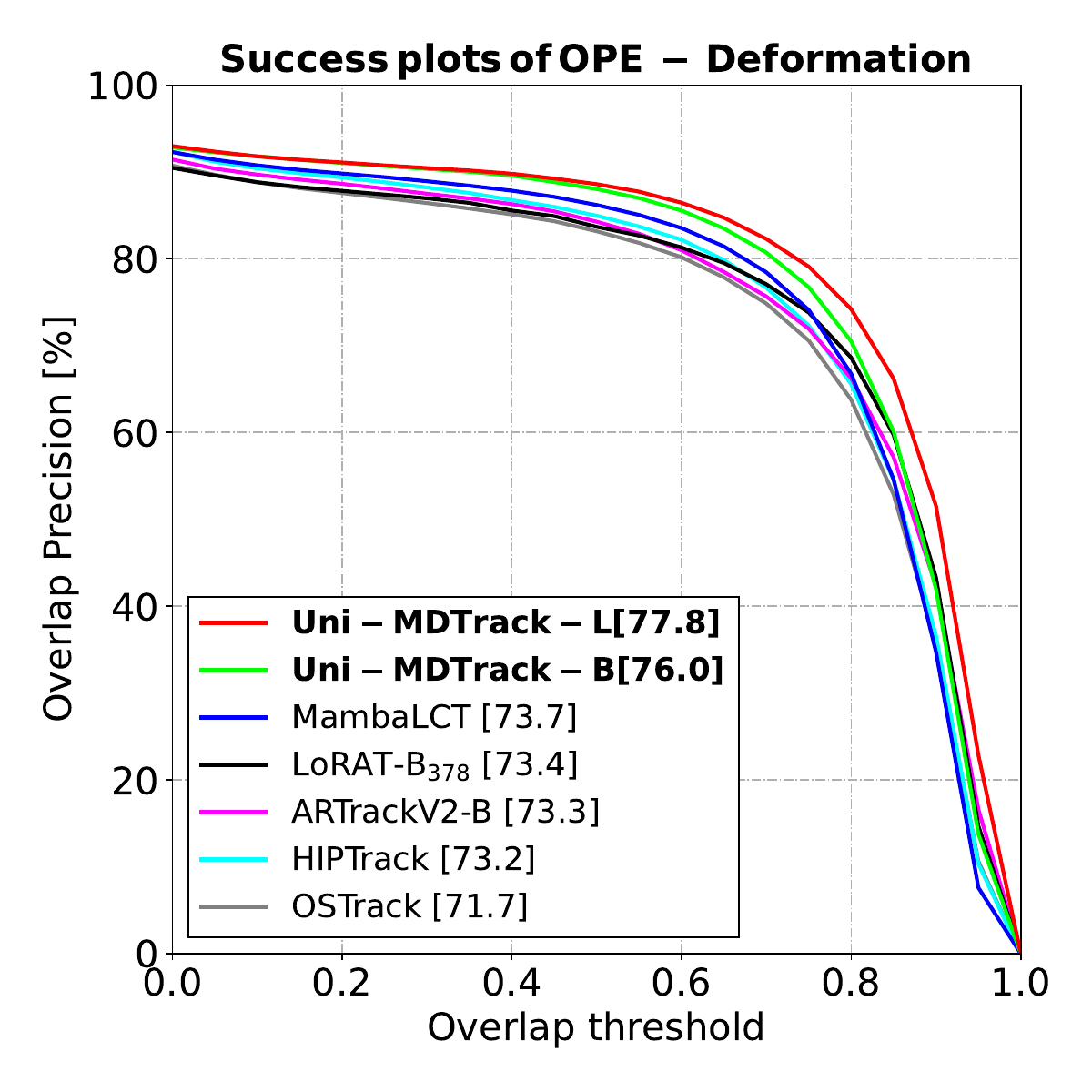}
\end{minipage}
}
{
\begin{minipage}{4.1cm}
\centering
\includegraphics[scale=0.23]{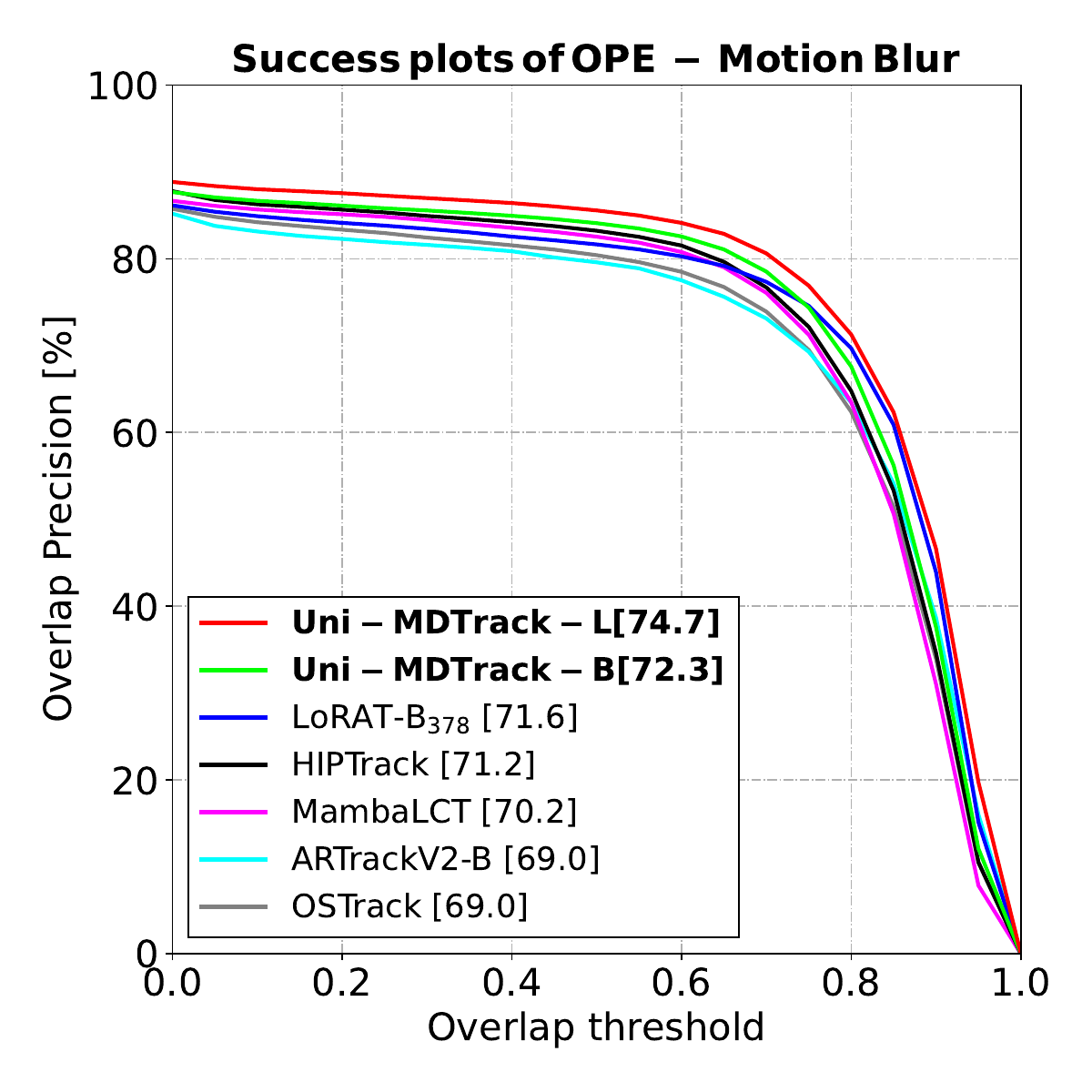}
\end{minipage}
}


{
\begin{minipage}{4.1cm}
\centering
\includegraphics[scale=0.23]{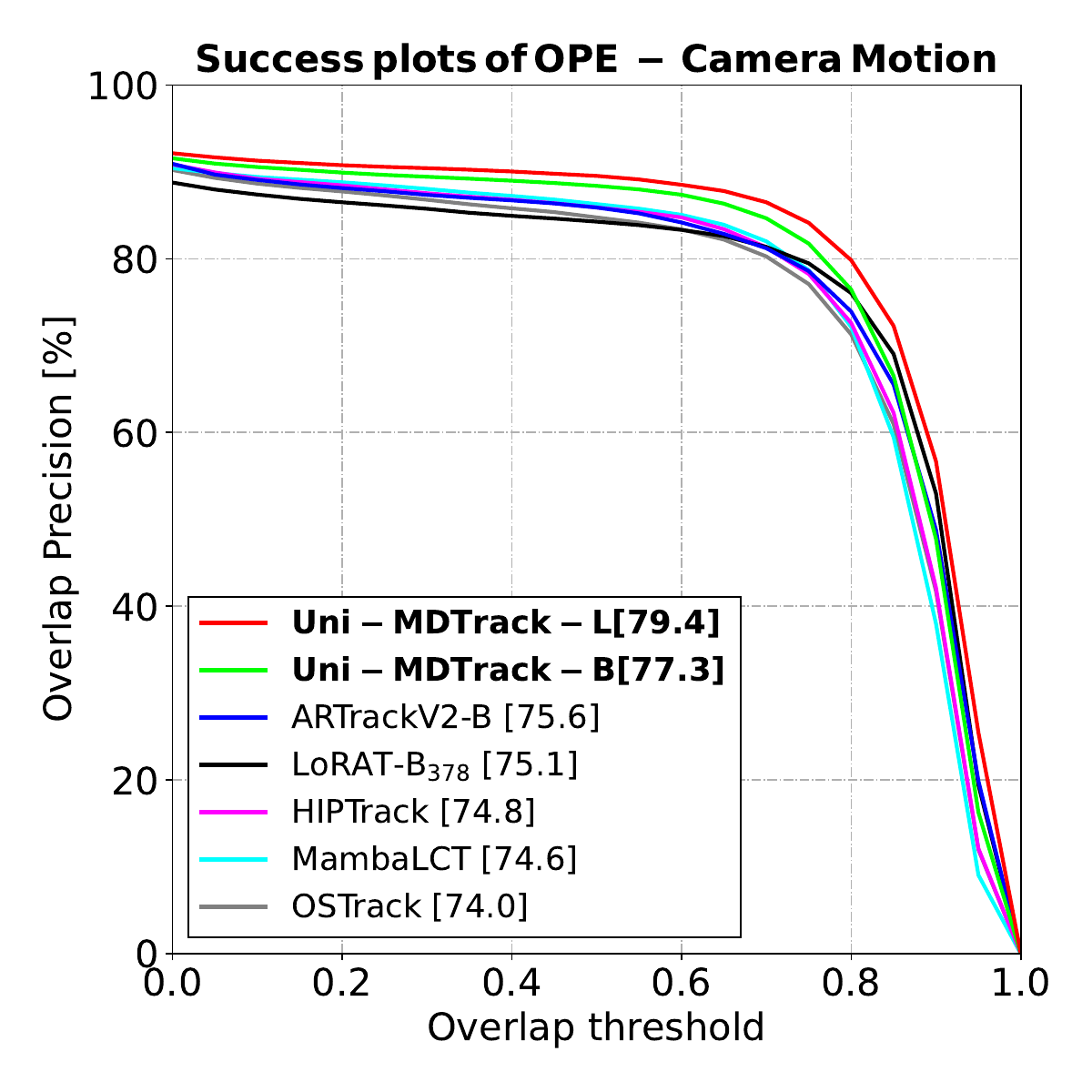}
\end{minipage}
}
{
\begin{minipage}{4.1cm}
\centering
\includegraphics[scale=0.23]{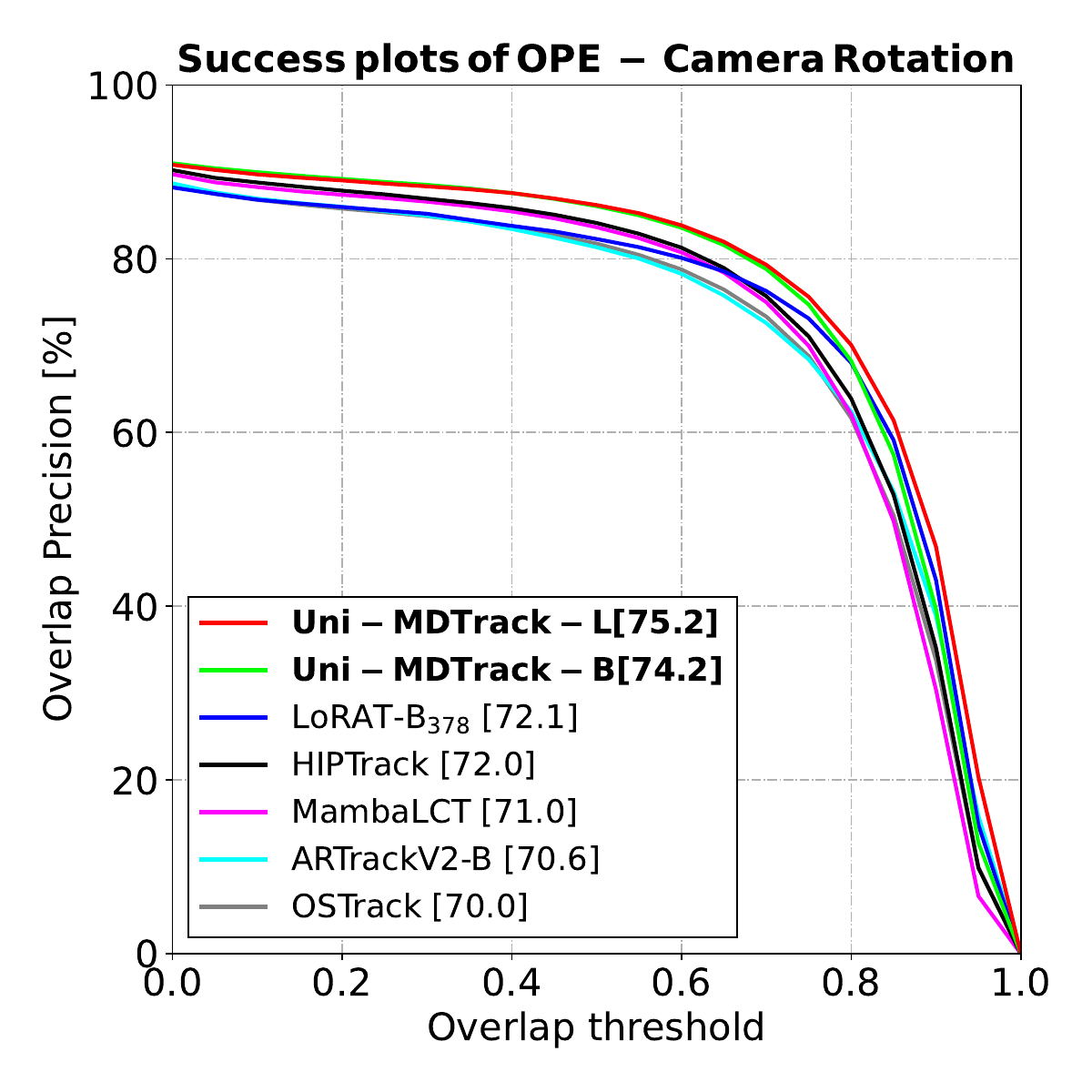}
\end{minipage}
}
{
\begin{minipage}{4.1cm}
\centering
\includegraphics[scale=0.23]{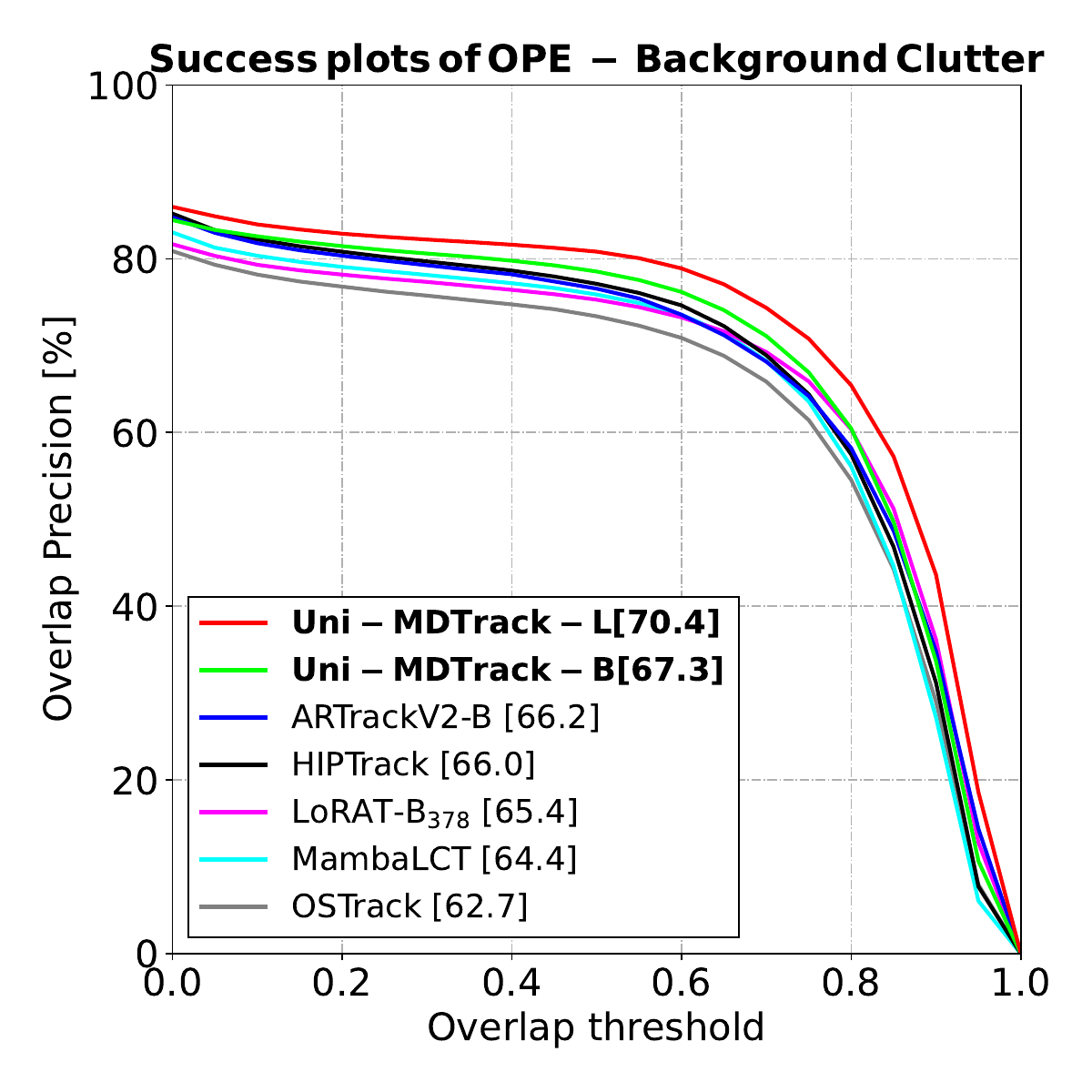}
\end{minipage}
}
{
\begin{minipage}{4.1cm}
\centering
\includegraphics[scale=0.23]{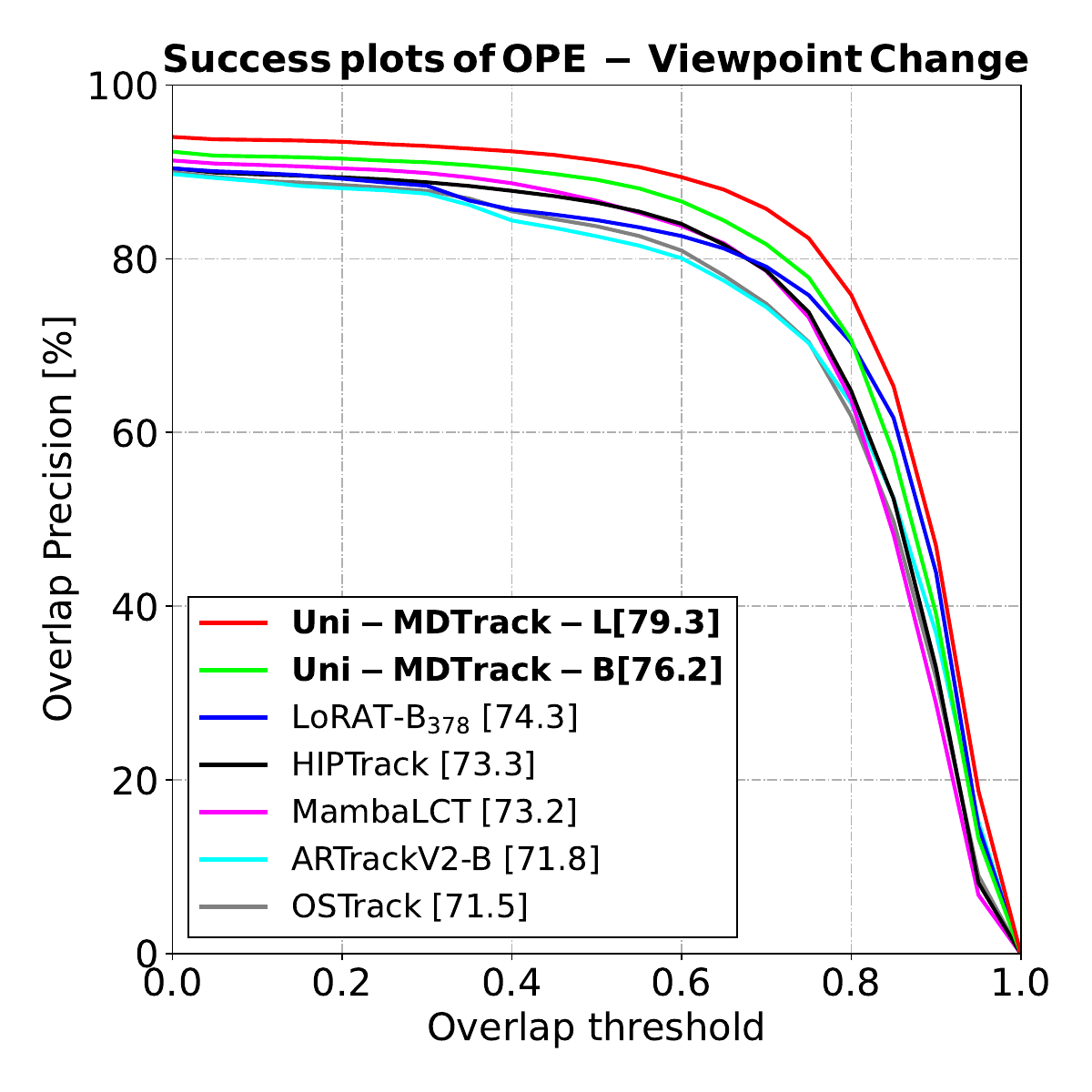}
\end{minipage}
}


{
\begin{minipage}{4.1cm}
\centering
\includegraphics[scale=0.23]{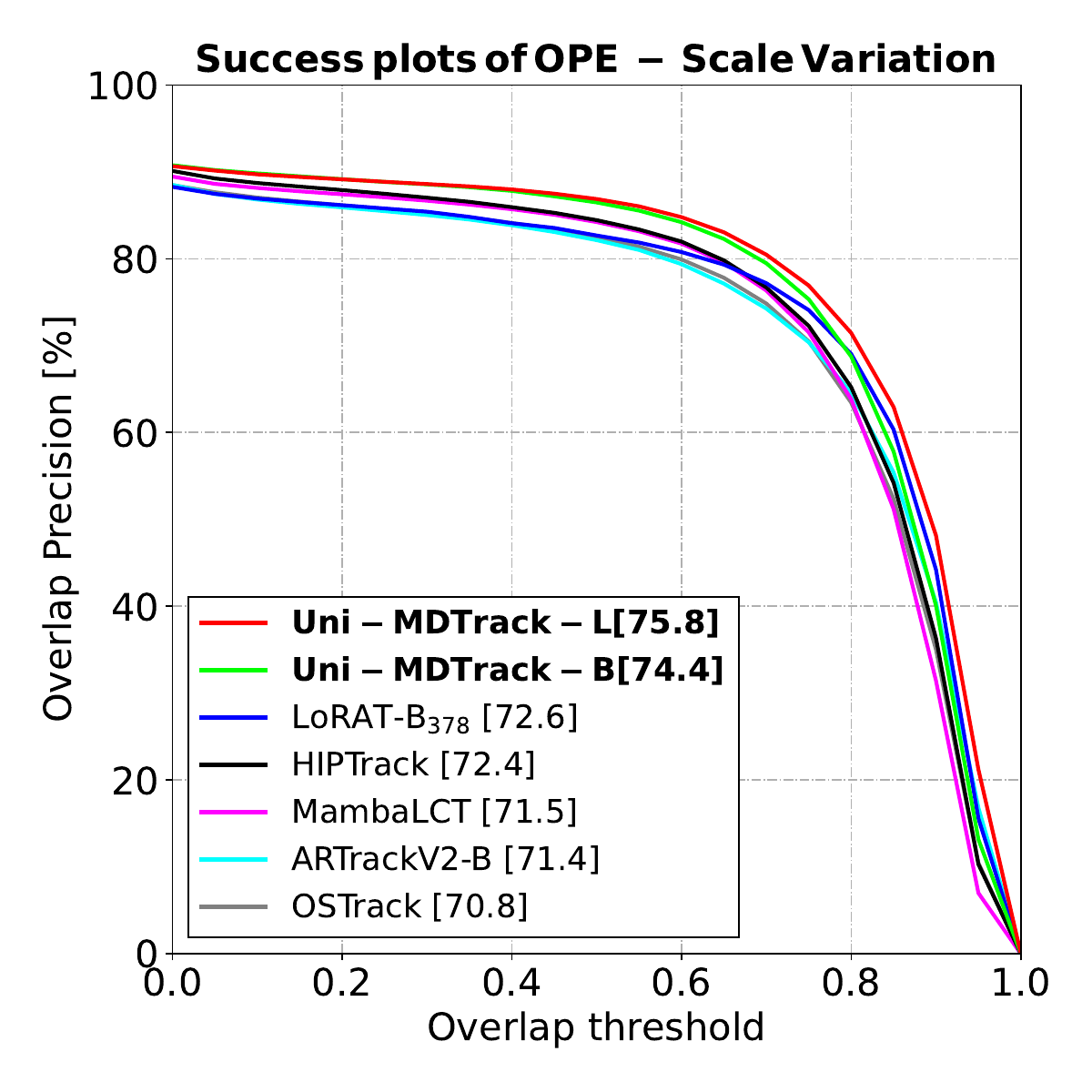}
\end{minipage}
}
{
\begin{minipage}{4.1cm}
\centering
\includegraphics[scale=0.23]{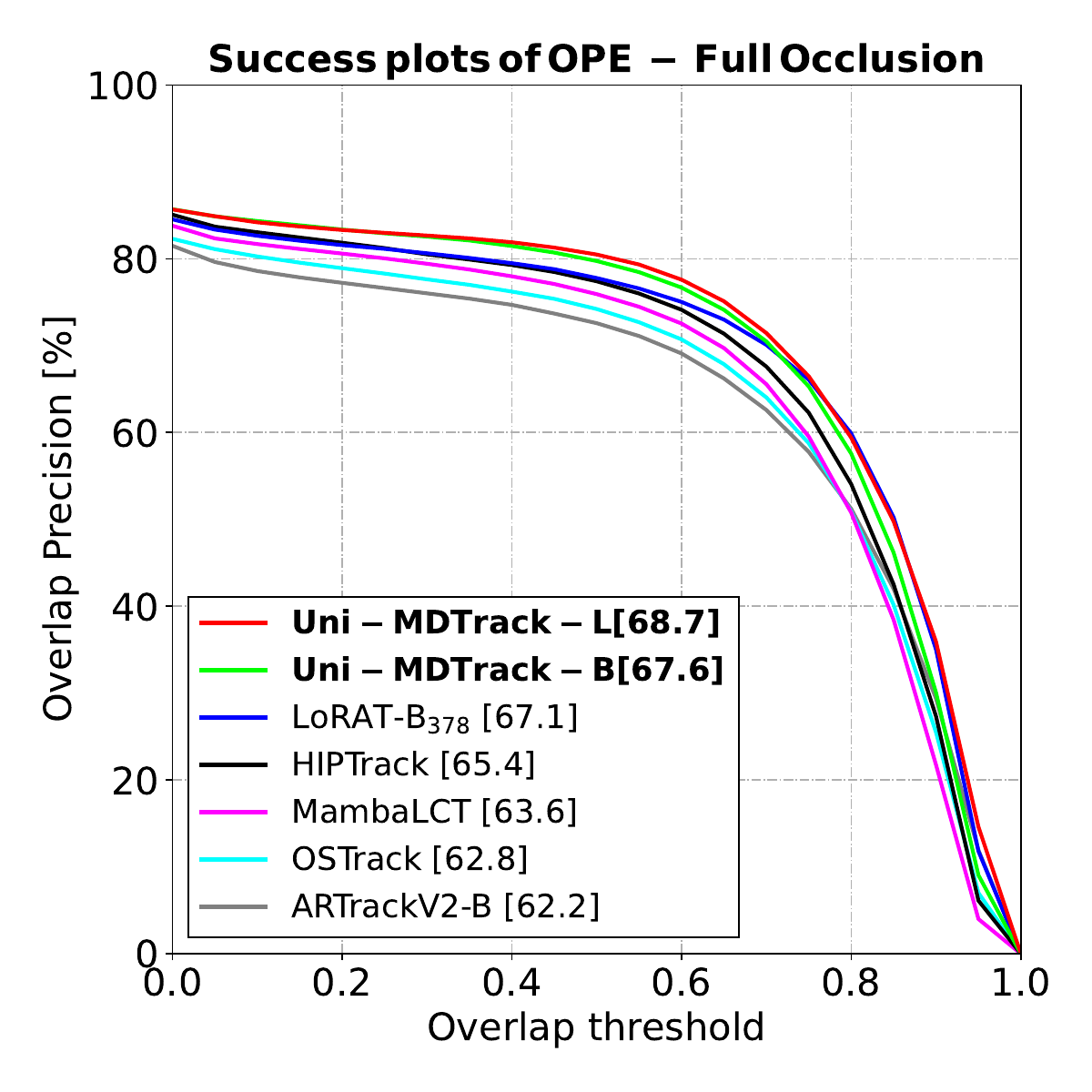}
\end{minipage}
}
{
\begin{minipage}{4.1cm}
\centering
\includegraphics[scale=0.23]{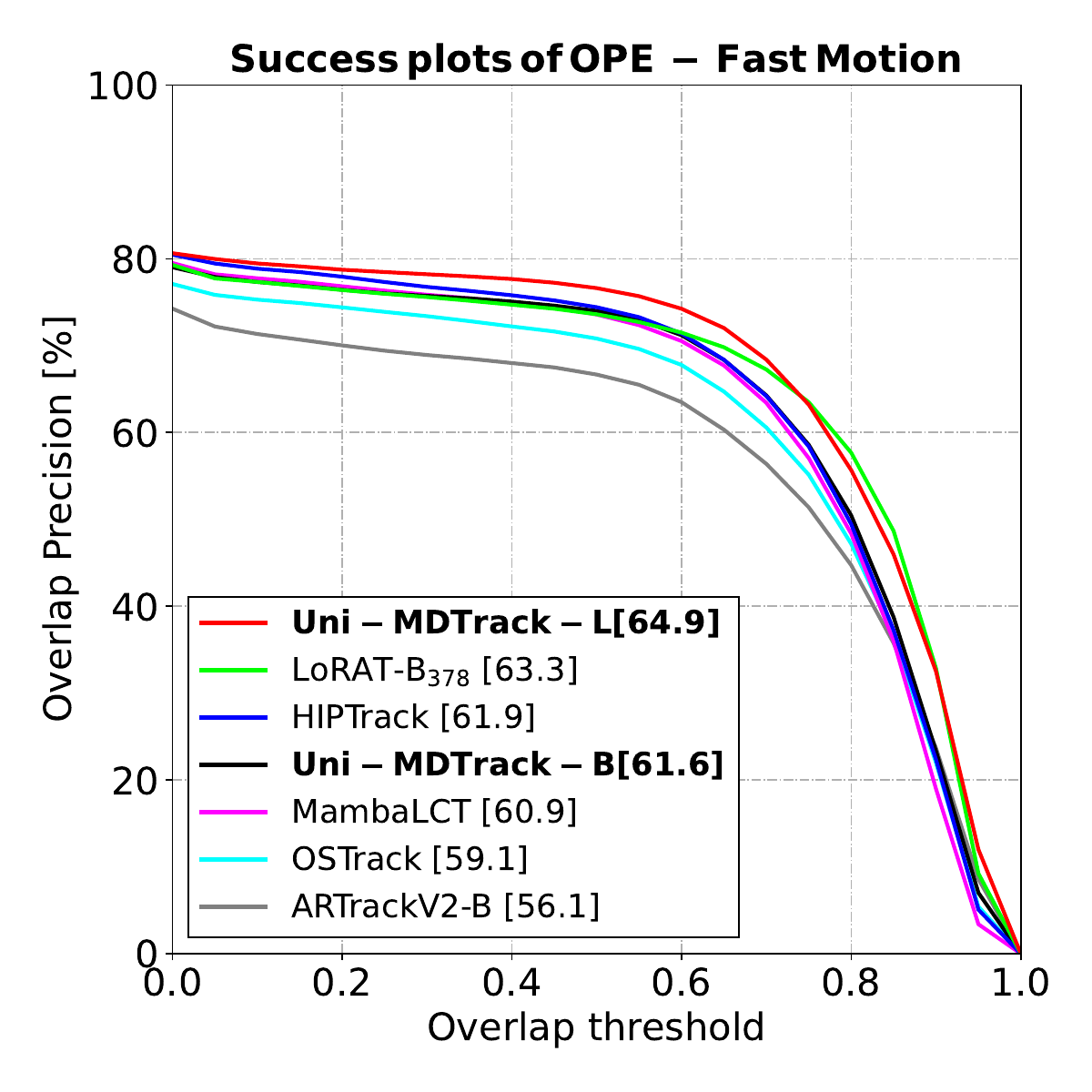}
\end{minipage}
}
{
\begin{minipage}{4.1cm}
\centering
\includegraphics[scale=0.23]{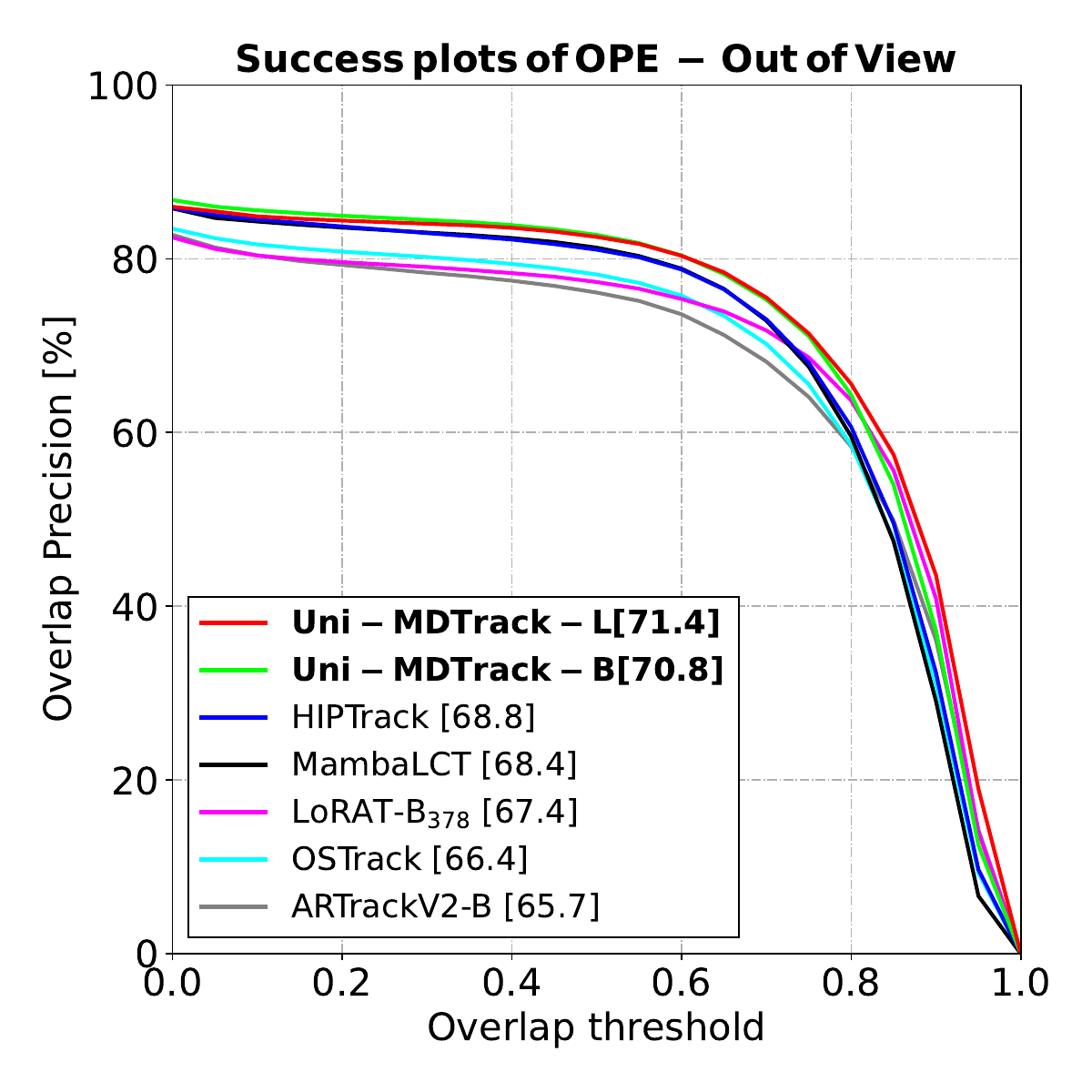}
\end{minipage}


{
\begin{minipage}{4.1cm}
\centering
\includegraphics[scale=0.23]{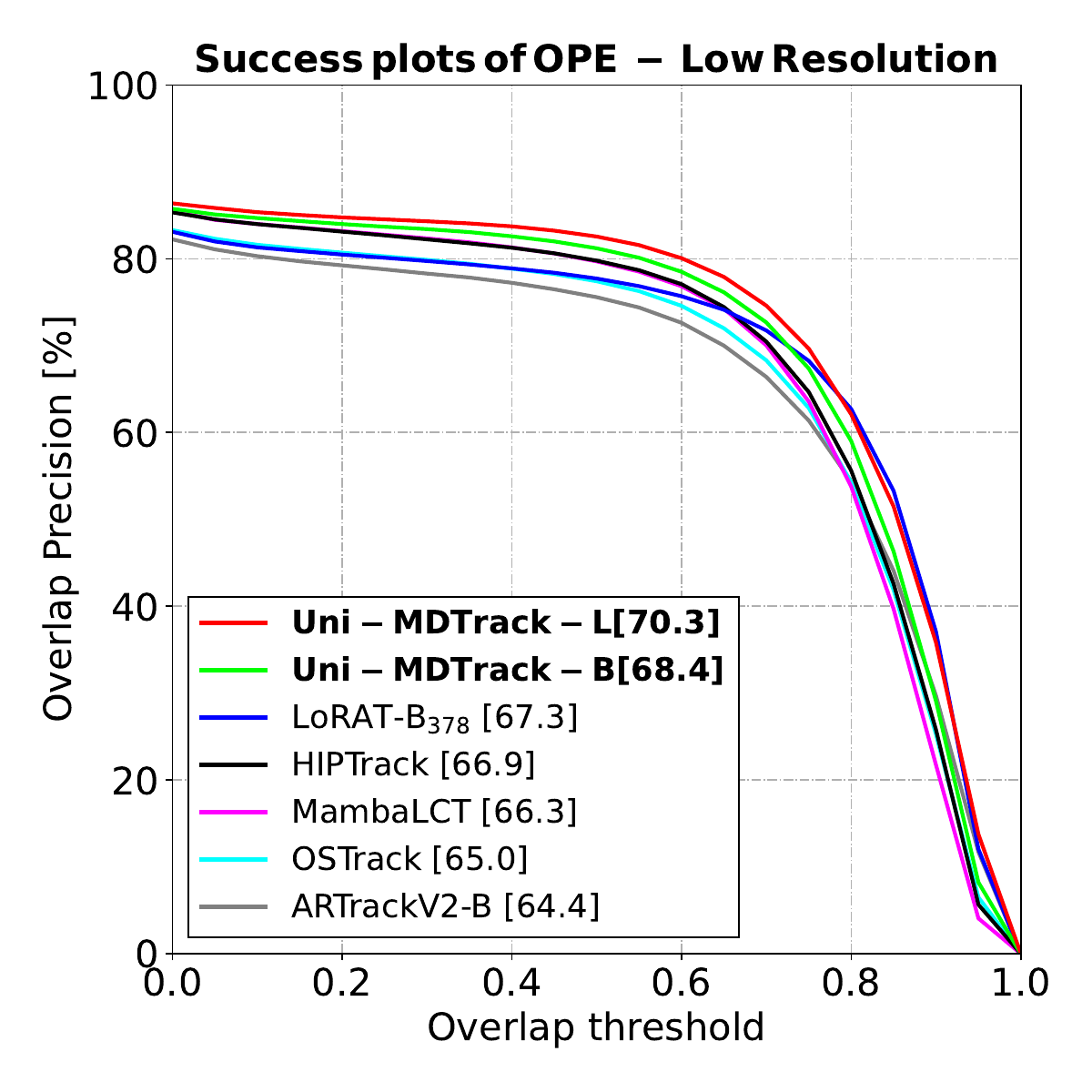}
\end{minipage}
}
{
\begin{minipage}{4.1cm}
\centering
\includegraphics[scale=0.23]{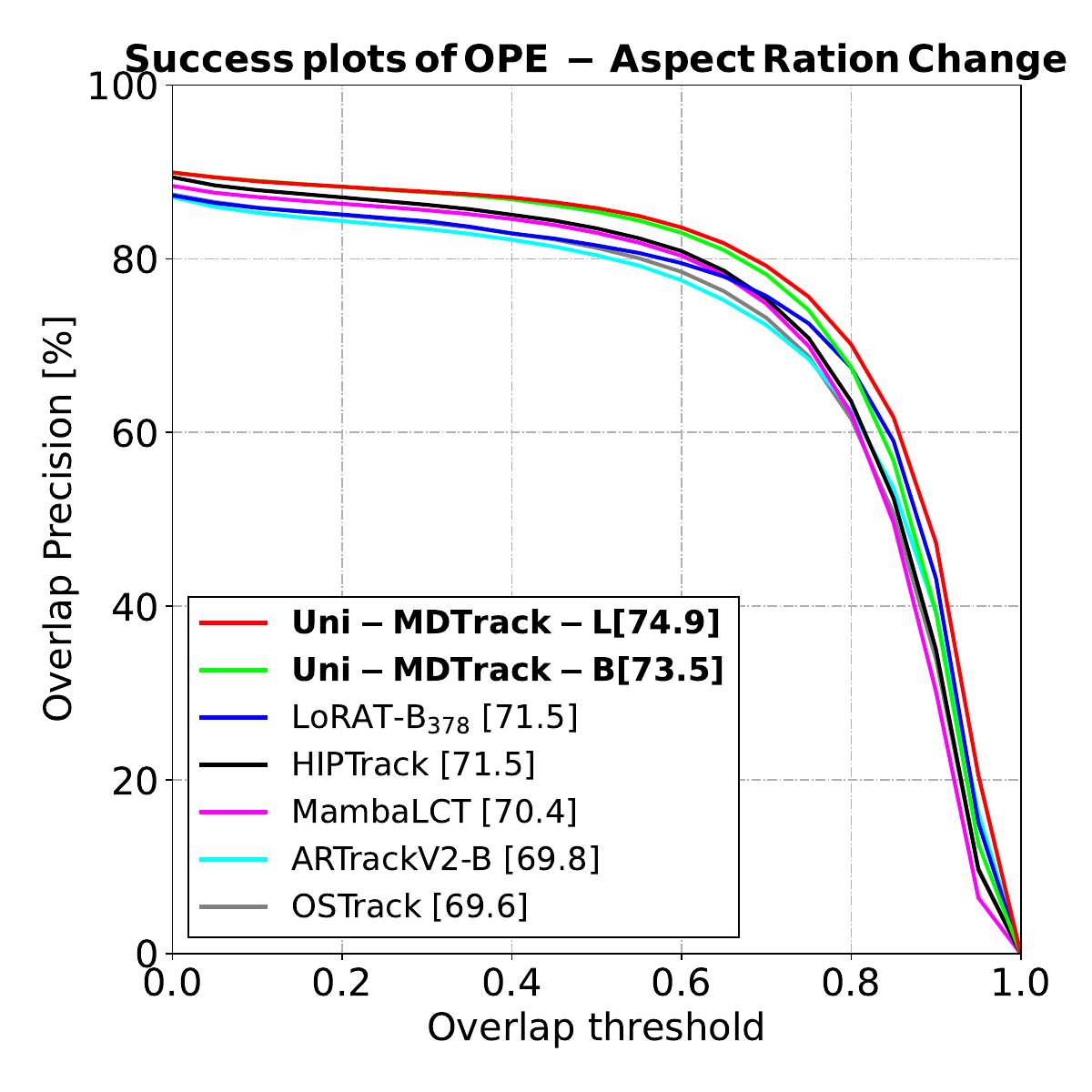}
\end{minipage}
}
{
\begin{minipage}{4.1cm}
\centering
\includegraphics[scale=0.23]{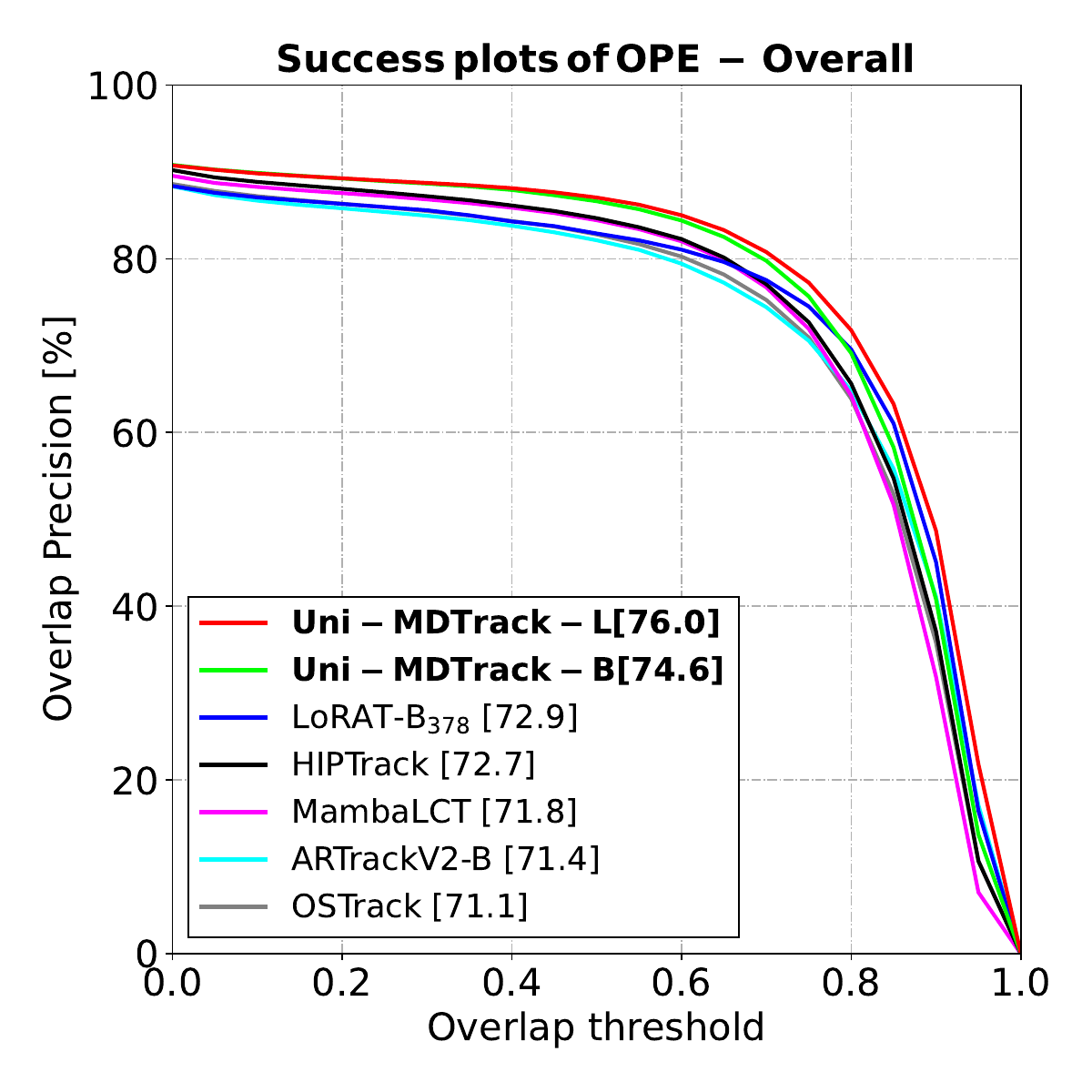}
\end{minipage}
}
{
\begin{minipage}{4.1cm}
\centering
\includegraphics[scale=0.23]{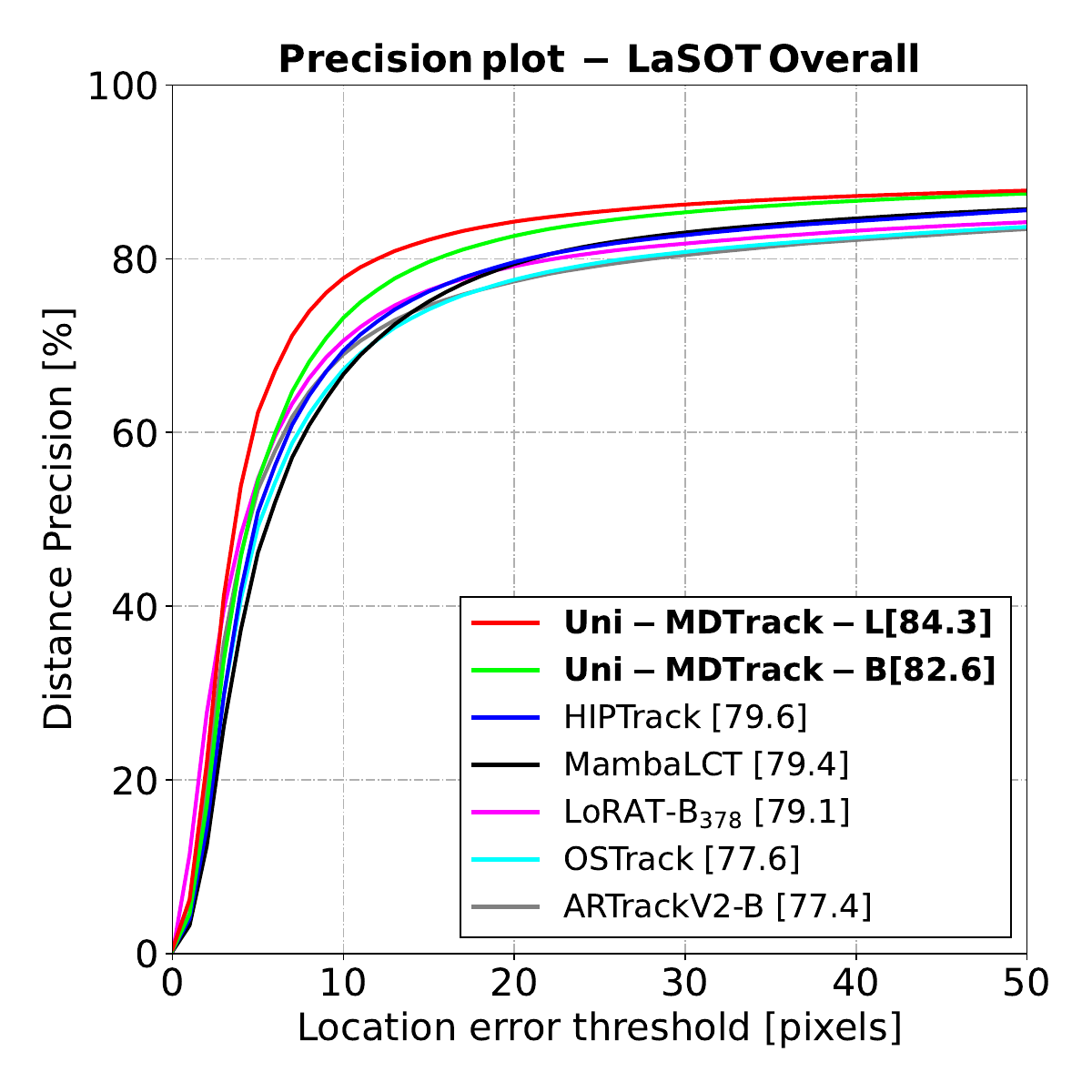}
\end{minipage}
}
}
\caption{Comparisons of our proposed Uni-MDTrack with other excellent trackers in the success curve on LaSOT \emph{test} split, which includes eleven challenging scenarios such as Low Resolution, Motion Blur, Scale Variation, etc. We also provide the comparisons of the success and precision curves across the entire LaSOT \emph{test} split. Zoom in for better view.}
\label{fig:comparision of lasot dataset}
\vspace{-1ex}
\end{figure*}

LaSOT~\cite{fan2019lasot} is well-known for featuring a diverse range of challenging tracking scenarios, therefore, 
in Figure \ref{fig:comparision of lasot dataset}, we provide a more detailed comparison of our proposed Uni-MDTrack-B with other current excellent trackers MambaLCT$_{256}$~\cite{li2025mambalct}, LoRAT-B$_{378}$~\cite{LoRAT}, HIPTrack \cite{Cai_2024_CVPR_HIPTrack}, ARTrackV2 \cite{Bai_2024_CVPR_ARTrackV2}, and OSTrack \cite{ye_2022_joint} across various challenging scenario subsets in LaSOT \cite{fan2019lasot}.
Figure \ref{fig:comparision of lasot dataset} presents detailed success curves and AUC scores across individual subsets, along with the success and precision curves on the entire LaSOT \emph{test} split. The results demonstrate that our Uni-MDTrack significantly outperforms these RGB-based trackers both overall and across the vast majority of subsets.

\subsection{Ablation Study}

 \noindent\textbf{The Importance of MCP and DSF.} In Table \ref{table_mcp_dsf}, based on Uni-MDTrack-B, we conduct ablation studies on the two core components of our method: MCP and DSF. Results demonstrates that both our proposed MCP and DSF modules individually contribute to significant tracking performance improvements. 
 MCP proves more critical for long-term (LaSOT) and infrared (LasHeR) tracking, whereas DSF generally has a greater impact on the remaining datasets.

  \begin{table}[h]\small
    \centering
    \caption{Ablation studies on MCP and DSF modules. Experiments are conducted on LaSOT (evaluated by AUC), LasHeR (SR), VisEvent (AUC) and DepthTrack (F-Score). }
    \resizebox{1.0\linewidth}{!}{%
    \begin{tabular}{c|cc|cccc|c}
    \Xhline{2pt}
        \textbf{\#} & \textbf{MCP} & \textbf{DSF}  &  LaSOT & LasHeR & VisEvent & DepthTrack & $\bm{\Delta}$ \\
        \Xhline{1pt}
        \textbf{1} & \ding{56} & \ding{56} & 73.2 & 59.9 & 62.7 & 65.1 & 0 \\ 
        \textbf{2} & \ding{56} & \ding{52} & 73.8 & 60.4 & 63.6 & 65.7 & \textbf{+0.65} \\
        \textbf{3} & \ding{52} & \ding{56} & 74.1 & 60.6 & 63.3 & 65.3 & \textbf{+0.6} \\
        \textbf{4} & \ding{52} & \ding{52} & 74.7 & 61.2 & 64.2 & 65.9 & \textbf{+1.3} \\
        
    \Xhline{2pt}
    \end{tabular}
    }
    \label{table_mcp_dsf}
\end{table}

\begin{table}[h]\small
    \centering
    \caption{A performance comparison of existing trackers and their integration with our method on LaSOT \emph{test} set.}
    \resizebox{1.0\linewidth}{!}{%
    \begin{tabular}{c|ccc}
    \Xhline{2pt}
        \textbf{Method} & AUC(\%) & $P_{Norm}$(\%) & $P$(\%)  \\
        \Xhline{1pt}
        DropTrack  & 71.8 & 81.8 & 78.1 \\
        DropTrack \textit{w/} \textbf{Ours} & \textbf{73.1} & 82.7 & \textbf{79.7}\\
        DropTrack \textit{w/} HIP & 72.7 & \textbf{82.9} & 79.5 \\
        DropTrack \textit{w/} Temporal Token & 72.0 & 81.9 & 78.4  \\
    \Xhline{2pt}
    \end{tabular}
    }
    \label{table_Generalization}
\end{table}

\noindent \textbf{Comparison with Other PEFT Methods.}
Table \ref{table_Generalization} presents a comparative analysis of our method against two prominent PEFT methods: HIPTrack and SPMTrack. To ensure a fair and controlled comparison, all methods are trained on the same foundation model DropTrack~\cite{Wu_2023_CVPR_dropmae}, which is a variant of the well-established OSTrack~\cite{ye_2022_joint}, featuring an enhanced initialization strategy.
Our method achieved a more significant performance improvement with only 50 epochs of training.

 \noindent \textbf{Generalization Ability of Our Method. }
 The results across Tables \ref{whole_comparison_dte}, \ref{whole_comparison}, and \ref{table_Generalization} collectively demonstrate the remarkable generalization ability and effectiveness of our proposed MCP and DSF modules. Our method consistently delivers substantial performance gains when applied to three distinct excellent trackers, proving performance gain in both pure RGB and unified multi-modal tracking scenarios.

 \begin{table}[!h]\small
    \centering
    \caption{Ablation study on different number of memory-aware compression tokens.}
    \setlength{\tabcolsep}{4.3mm}
    \begin{tabular}{c|cccc}
    \Xhline{2pt}
        \textbf{Number}  & \textbf{8} & \textbf{16} & \textbf{32} & \textbf{64} \\
        \Xhline{1pt}
        LaSOT & 74.2 & 74.7 & 74.7 & 74.6 \\
        LasHeR & 60.5 & 61.2 & 61.3 & 61.3 \\
        VisEvent & 63.8 & 64.2 & 64.2 & 64.3\\
        DepthTrack & 65.5 & 65.9 & 65.9 & 66.1 \\
        $\bm{\Delta}$ & \textbf{-0.5} & \textbf{0} & \textbf{+0.03} & \textbf{+0.08} \\
        
    \Xhline{2pt}
    \end{tabular}
    \label{table3_number_of_tokens}
\end{table}

\begin{figure*}[!ht]
  \centering
    \subfloat[Qualitative results of three methods when the targets in large scale variations.]{\includegraphics[width=\textwidth]{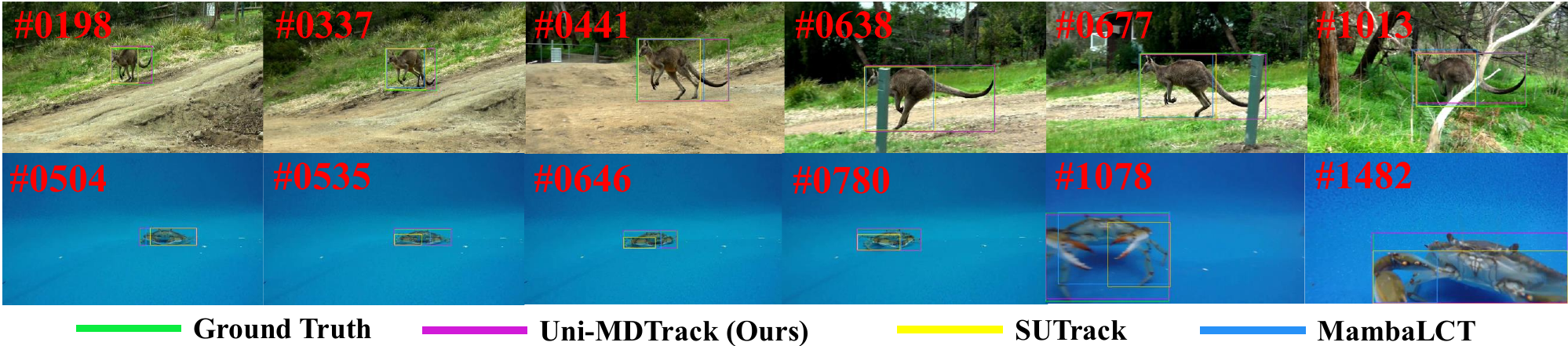}\label{fig-vis-1}} \\
    \subfloat[ Qualitative results of three methods when the targets are among similar objects and suffer partial occlusion.]
    {\includegraphics[width=\textwidth]{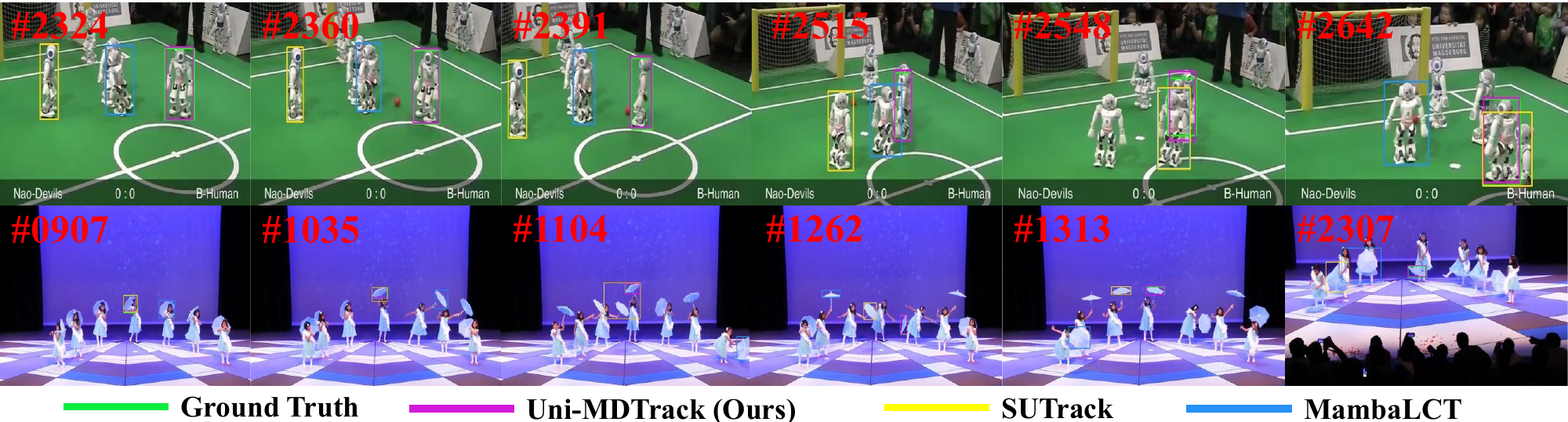}\label{fig-vis-2}} \\
    \subfloat[ Qualitative results of three methods when the targets have large occlusions and sudden moves.]{\includegraphics[width=\textwidth]{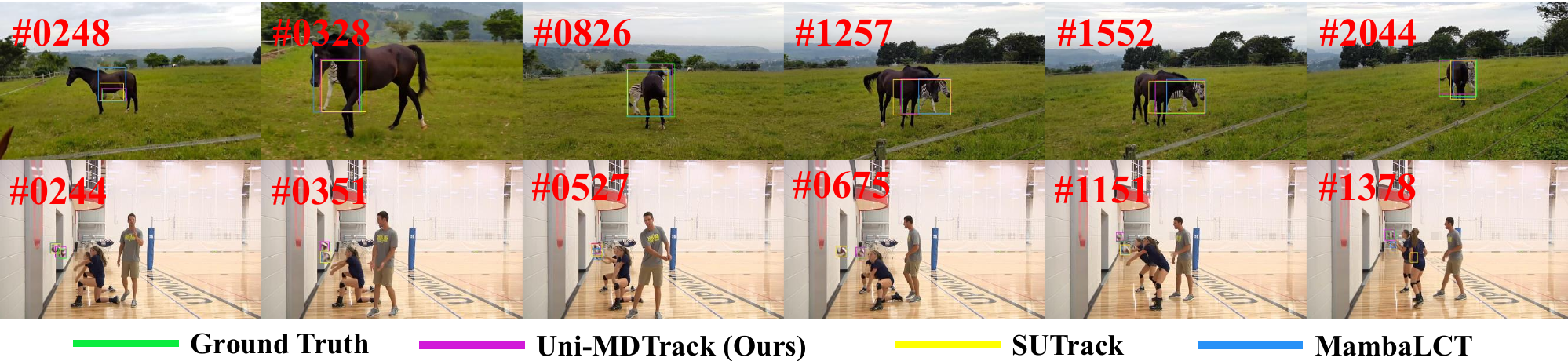}\label{fig-vis-3}} \\
  \caption{ This figure presents a visual comparison among our proposed Uni-MDTrack, MambaLCT$_{256}$ \cite{li2025mambalct} and SUTrack-B \cite{sutrack} in the challenges of target among similar objects, undergoes sudden movements, partial occlusion and scale variation. It demonstrates that our method achieves more effective and accurate tracking in the aforementioned challenging scenarios. Zoom in for better view.}
    \label{fig:qualitative}
\end{figure*}

\begin{figure*}[!t]
    \centering
    \includegraphics[width=\textwidth]{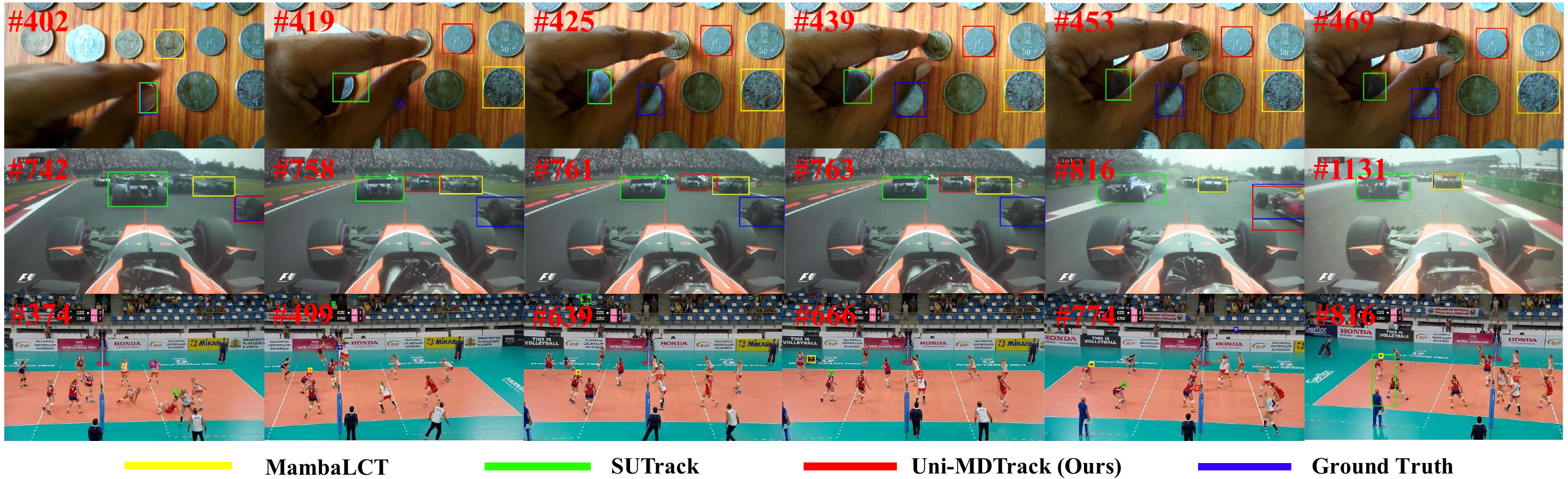}
    \caption{
    This figure illustrates cases of tracking failure under extremely complex scenarios. When the target undergoes severe occlusion and fast motion, while heavy background clutter are present, all methods experience tracking failures. Zoom in for better view.}
    \label{fig:vis_bascase}
\end{figure*}

 \noindent \textbf{The Number of Memory-Aware Compression Token.}
 Table \ref{table3_number_of_tokens} explores the impact of the number of memory-aware compression tokens based on Uni-MDTrack-B. While performance increases with more tokens, it eventually saturates. We thus selected 16 as a balanced choice.

 \noindent \textbf{The Number of DSF Modules.}
Our standard model employs four DSF modules, with each DSF module uniformly fusing target dynamic state features into the input and output of certain layers in the backbone network. In Table \ref{table3_number_of_dsf_base}, we attempted to introduce different numbers of DSF modules based on Uni-MDTrack-B, respectively. We still divided its last 24 layers into equal parts.
The results show that using 4 DSF modules achieves the best performance; more DSF modules will introduce additional parameters, and 50 epochs of training may not be sufficient to fully train these modules. Meanwhile, due to time constraints, we do not conduct further experiments on Uni-MDTrack-L, so we adopt the same settings as Uni-MDTrack-B.

\begin{table}[!h]\small
    \centering
    \caption{Ablation study on different number of DSF modules based on Uni-MDTrack-B.}
    \setlength{\tabcolsep}{4.3mm}
    \begin{tabular}{c|cccc}
    \Xhline{2pt}
        \textbf{Number}  & \textbf{2} & \textbf{4} & \textbf{6} & \textbf{8} \\
        \Xhline{1pt}
        LaSOT & 74.3 & 74.7 & 74.7 & 74.6 \\
        LasHeR & 61.0 & 61.2 & 61.4 & 61.2 \\
        VisEvent & 63.6 & 64.2 & 64.2 & 64.0 \\
        DepthTrack & 65.4 & 65.9 & 65.9 & 65.7 \\
        $\bm{\Delta}$ & \textbf{-0.43} & \textbf{0} & \textbf{+0.05} & \textbf{-0.13} \\
        
    \Xhline{2pt}
    \end{tabular}
    \label{table3_number_of_dsf_base}
\end{table}

\noindent \textbf{Which Features to Use for Target State Updates.}
A core design principle of our method is to update the state of all DSF modules using only the final search region features. Isolating the state update from template information ensures that the DSF modules are solely dedicated to capturing the real-time dynamics of the target. Table \ref{table_dsf_update} presents an ablation study where we validate this design choice by investigating the impact of using different feature sources for the state update.
The second row of Table \ref{table_dsf_update} represents using the overall output sequence to perform SSM state updates; the third row represents using the search region features corresponding to the input of the DSF module's input fusion layer to perform updates, \textit{i.e.}, the input for state updates of each DSF module comes from different intermediate layers of the backbone.  The results show that introducing other features reduces model performance; meanwhile, using intermediate network layer features to update the target state is not as effective as using the final search region features.

\begin{table}[t]\small
    \centering
    \caption{Ablation studies on which features to use for target state updates in DSF.}
    \resizebox{1.0\linewidth}{!}{%
    \begin{tabular}{c|c|cccc|c}
    \Xhline{2pt}
        \textbf{\#} & \textbf{Variants}  &  LaSOT & LasHeR & VisEvent & DepthTrack & $\bm{\Delta}$ \\
        \Xhline{1pt}
        \textbf{1} & Output Search Feature & 74.7 & 61.2 & 64.2 & 65.9 & 0  \\ 
        \textbf{2} & Output Whole Sequence & 74.5 & 60.9 & 63.9 & 65.7 & \textbf{-0.25}  \\
        \textbf{3} & Input Fusion Sequence & 74.5 & 60.7 & 64.1 & 65.6 & \textbf{-0.28} \\
        
    \Xhline{2pt}
    \end{tabular}
    }
    \label{table_dsf_update}
    \vspace{-2ex}
\end{table}

\noindent \textbf{The Impact of Using State Space Model.}
Our DSF module employs a Mamba-like SSM~\cite{mamba}. In fact, in addition to state space models, there are many RNN-like structures that can achieve state updates. As shown in Table \ref{table_SSM}, we also experimented with replacing SSM with a simple LSTM. The results show that SSM can achieve better performance due to better state update algorithms, but LSTM is still effective. The more important contribution of the DSF module lies in demonstrating that continuous target state change features can serve as an effective complement to the backbone, rather than the specific design of the module itself.

\noindent \textbf{Comparison with Using SSMs as Backbone Layers.}
Previous methods such as MambaVT~\cite{lai2025mambavt} and MambaVLT~\cite{Liu_2025_CVPR_mambavlt} predominantly use SSM as the backbone network implementation, or replace certain layers of the backbone with SSM, fundamentally playing the same role as other backbone layers while merely utilizing SSM's linear computational complexity to extend context length. These approaches also require designing complex scanning algorithms. But linear SSMs are not guaranteed to outperform Transformers under the same sequence length (as proven in ~\cite{10.5555/3692070.3693514_illusion_of_SSM}). We have conducted experiments demonstrating this limitation, where replacing our DSF module with the same number of encoder layers from MambaVT and inserting them at regular intervals throughout the backbone network, which results in substantial performance degradation, as shown in Table \ref{table_SSM_backbone}.

 \begin{table}[t]\small
    \centering
    \caption{Ablation studies on replacing SSM with other model.}
    \resizebox{1.0\linewidth}{!}{%
    \begin{tabular}{c|c|cccc|c}
    \Xhline{2pt}
        \textbf{\#} & \textbf{Variants}  &  LaSOT & LasHeR & VisEvent & DepthTrack & $\bm{\Delta}$ \\
        \Xhline{1pt}
        \textbf{1} & SSM &  74.7 & 61.2 & 64.2 & 65.9 & 0 \\ 
        \textbf{2} & LSTM &  74.4 & 60.8 & 63.8 & 65.7 & \textbf{-0.33} \\
        
    \Xhline{2pt}
    \end{tabular}
    }
    \label{table_SSM}
    \vspace{-2ex}
\end{table}

\noindent \textbf{The Size of Memory Bank.}
In this paper, we construct the memory bank by sampling the search region features from $n$ tracked frames, which are selected uniformly and at equal intervals from all previously tracked frames. By default, we set the number $n$ to 50. In Table \ref{table3_number_of_mcp}, we further experiment with different memory bank sizes.
When the memory bank size is further increased, model performance ceases to improve; thus, we select 50 to conserve GPU memory. This may be due to the limited number of queries and network layers of MCP, which restrict the compression capability for large memory bank.
Additionally, since DSF can capture the short-term dynamic state changes of the target, MCP is not required to sample more densely.

 \begin{table}[t]\small
    \centering
    \caption{Ablation studies on leveraging SSMs as backbone layers.}
    \resizebox{1.0\linewidth}{!}{%
    \begin{tabular}{c|c|cccc|c}
    \Xhline{2pt}
        \textbf{\#} & \textbf{Variants}  &  LaSOT & LasHeR & VisEvent & DepthTrack & $\bm{\Delta}$ \\
        \Xhline{1pt}
        \textbf{1} & \textbf{Ours} &  74.7 & 61.2 & 64.2 & 65.9 & 0 \\ 
        \textbf{2} & \textbf{Backbone Layers} &  74.0 & 60.5 & 63.5 & 65.4 & \textbf{-0.65} \\
        
    \Xhline{2pt}
    \end{tabular}
    }
    \label{table_SSM_backbone}
    \vspace{-2ex}
\end{table}

 \begin{table}[!h]\scriptsize
    \centering
    \caption{Ablation study on memory bank size of MCP module based on Uni-MDTrack-B.}
    \setlength{\tabcolsep}{5.4mm}
    \begin{tabular}{c|cccc}
    \Xhline{2pt}
        \textbf{Number}  & \textbf{10} & \textbf{20} & \textbf{50} & \textbf{100} \\
        \Xhline{1pt}
        LaSOT & 74.3 & 74.6 & 74.7 & 74.6\\
        LasHeR & 60.7 & 61.0 & 61.2  & 61.0 \\
        VisEvent & 63.9 & 64.0 & 64.2  & 64.4 \\
        DepthTrack & 65.9 & 66.0 & 65.9  & 65.8 \\
        $\bm{\Delta}$ & \textbf{-0.3} & \textbf{-0.1} & 0 & \textbf{-0.05} \\
        
    \Xhline{2pt}
    \end{tabular}
    \label{table3_number_of_mcp}
\end{table}

\begin{figure*}[!th]
  \centering
    \vspace{3ex}
    \subfloat[Qualitative results of three methods on RGB-Depth tasks.]{\includegraphics[width=\textwidth]{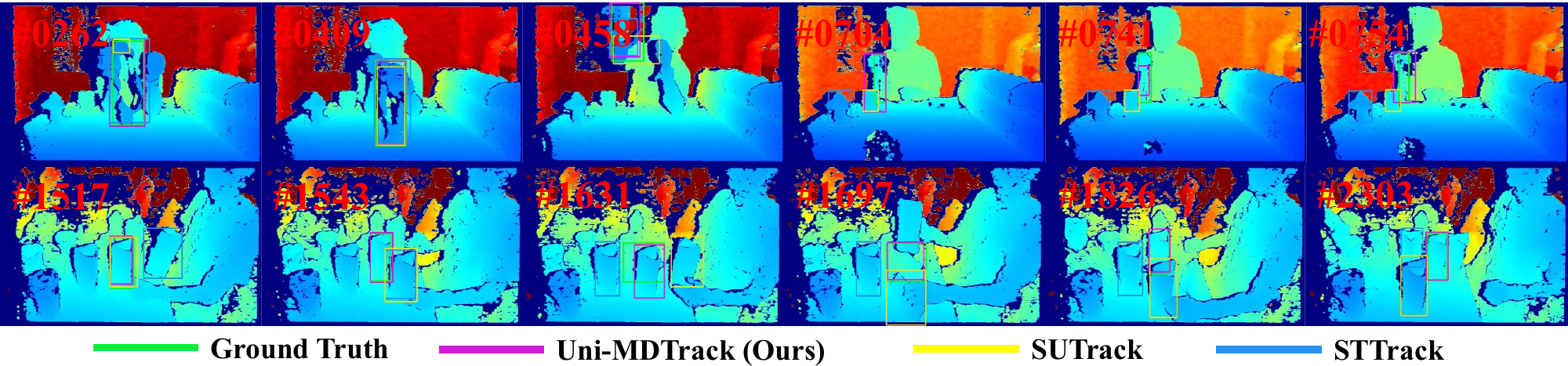}\label{fig-vis-1}} \\
    \subfloat[ Qualitative results of three methods on RGB-Event tasks. ]
    {\includegraphics[width=\textwidth]{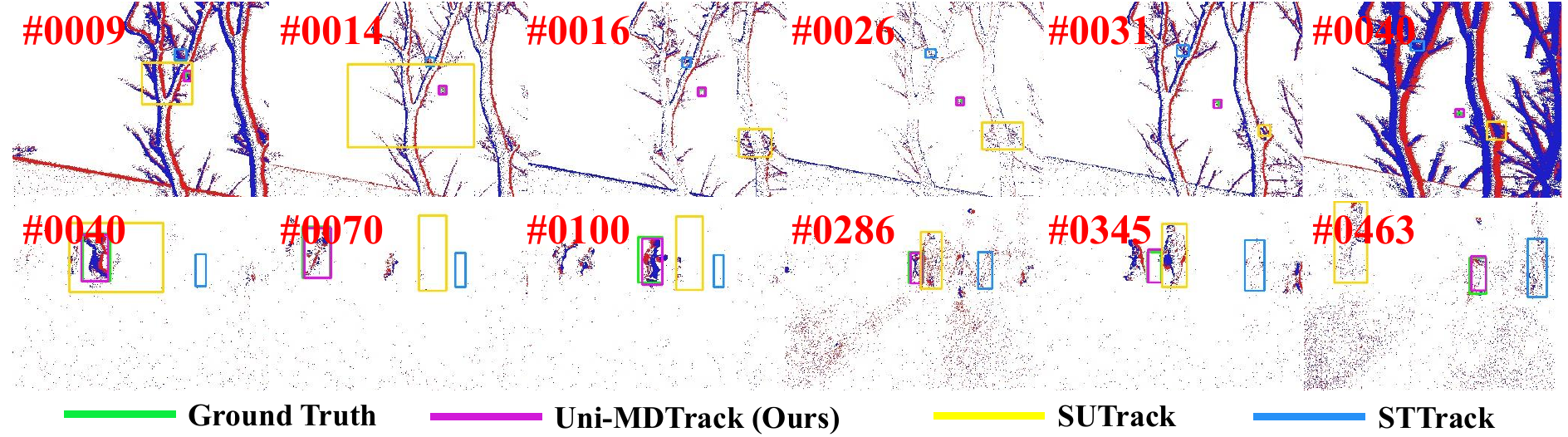}\label{fig-vis-2}} \\
    \subfloat[ Qualitative results of three methods on RGB-Thermal tasks.]{\includegraphics[width=\textwidth]{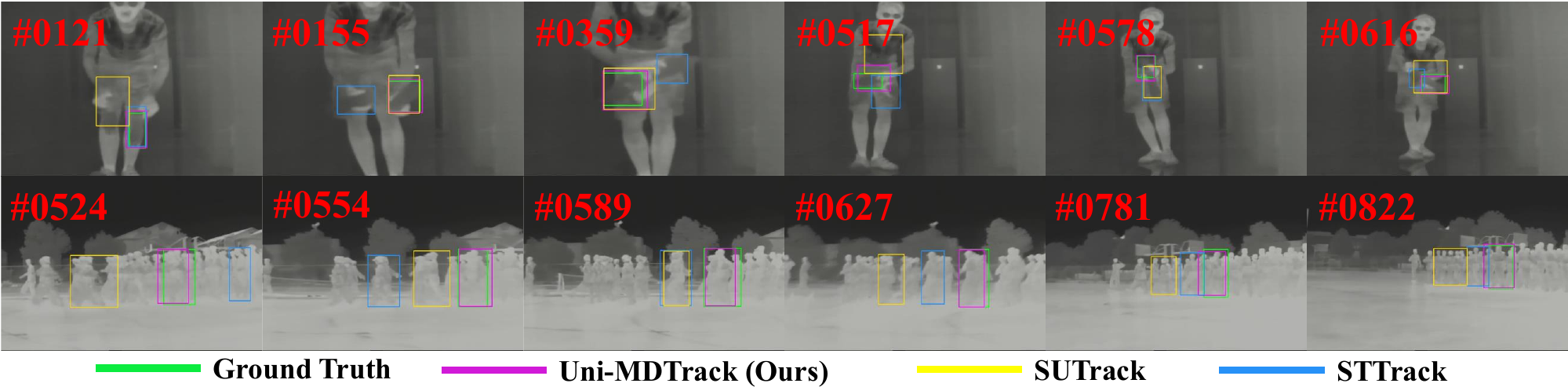}\label{fig-vis-3}} \\
  \caption{ This figure presents a visual comparison among our proposed Uni-MDTrack-B,  STTrack \cite{hu2025exploiting_sttrack} and SUTrack-B \cite{sutrack} in the challenges of different modalities including RGB-Depth, RGB-Event and RGB-Thermal tasks. It demonstrates that our method achieves more effective and accurate tracking in the aforementioned challenging scenarios. Zoom in for better view.}
    \label{fig:quantitative2}
\end{figure*}

\noindent \textbf{The Impact of Position Bias Added to Attention Score.}
When using queries for memory feature querying and aggregation, we incorporate the ALiBi~\cite{presstrain_alibi} positional bias for different memory frames to maintain the positional awareness of memory frames, enabling the model to pay more attention to newer features when encountering similar historical memories. In Table \ref{table_position_bias}, we experimented with not using positional information and using absolute positional encoding. The results show that ALiBi performs the best, while absolute positional encoding performs the worst. This is because we only sample 5 frames as search frames during training, requiring the introduction of positional encoding with extrapolation capability.

 \begin{table}[!h]\small
    \centering
    \caption{Ablation studies on position bias added to attention score in MCP.}
    \resizebox{1.0\linewidth}{!}{%
    \begin{tabular}{c|c|cccc|c}
    \Xhline{2pt}
        \textbf{\#} & \textbf{Variants}  &  LaSOT & LasHeR & VisEvent & DepthTrack & $\bm{\Delta}$ \\
        \Xhline{1pt}
        \textbf{1} & ALiBi &  74.7 & 61.2 & 64.2 & 65.9 & 0 \\ 
        \textbf{2} & \textit{w/o} Position & 74.6 & 61.2 & 64.0 & 65.9 & \textbf{-0.08} \\
        \textbf{3} & Absolute Position Encoding & 74.4 & 61.0 & 64.1 & 65.9 & \textbf{-0.15} \\
        
    \Xhline{2pt}
    \end{tabular}
    }
    \label{table_position_bias}
\end{table}

\noindent \textbf{Select Strategy of Memory.}
Our current memory bank selection strategy is to select n frames uniformly at equal intervals from all previously tracked frames. In addition, we also experimented with other selection strategies, with the results presented in Table \ref{table_mem_select}. We try adding search region features to the memory bank every 5 frames and 10 frames while using a first-in-first-out (FIFO) strategy to maintain the memory bank size.
The results show that uniform sampling can ensure longer-term memory and yields significant benefits, especially on long-term tracking datasets such as LaSOT.

 \begin{table}[!h]\small
    \centering
    \caption{Ablation studies on memory select strategy.}
    \resizebox{1.0\linewidth}{!}{%
    \begin{tabular}{c|c|cccc|c}
    \Xhline{2pt}
        \textbf{\#} & \textbf{Variants}  &  LaSOT & LasHeR & VisEvent & DepthTrack & $\bm{\Delta}$ \\
        \Xhline{1pt}
        \textbf{1} & \textbf{Ours} &  74.7 & 61.2 & 64.2 & 65.9 & 0 \\ 
        \textbf{2} & FIFO + 5 Interval & 74.1 & 61.1 & 63.8 & 65.8 & \textbf{-0.3}\\
        \textbf{3} & FIFO + 10 Interval & 74.2 & 61.1 & 63.9 &  65.9 & \textbf{-0.23} \\
        
    \Xhline{2pt}
    \end{tabular}
    }
    \label{table_mem_select}
    \vspace{-2ex}
\end{table}

\subsection{Qualitative Study}

\begin{figure}[!t]
    \centering
    \includegraphics[width=0.5\textwidth]{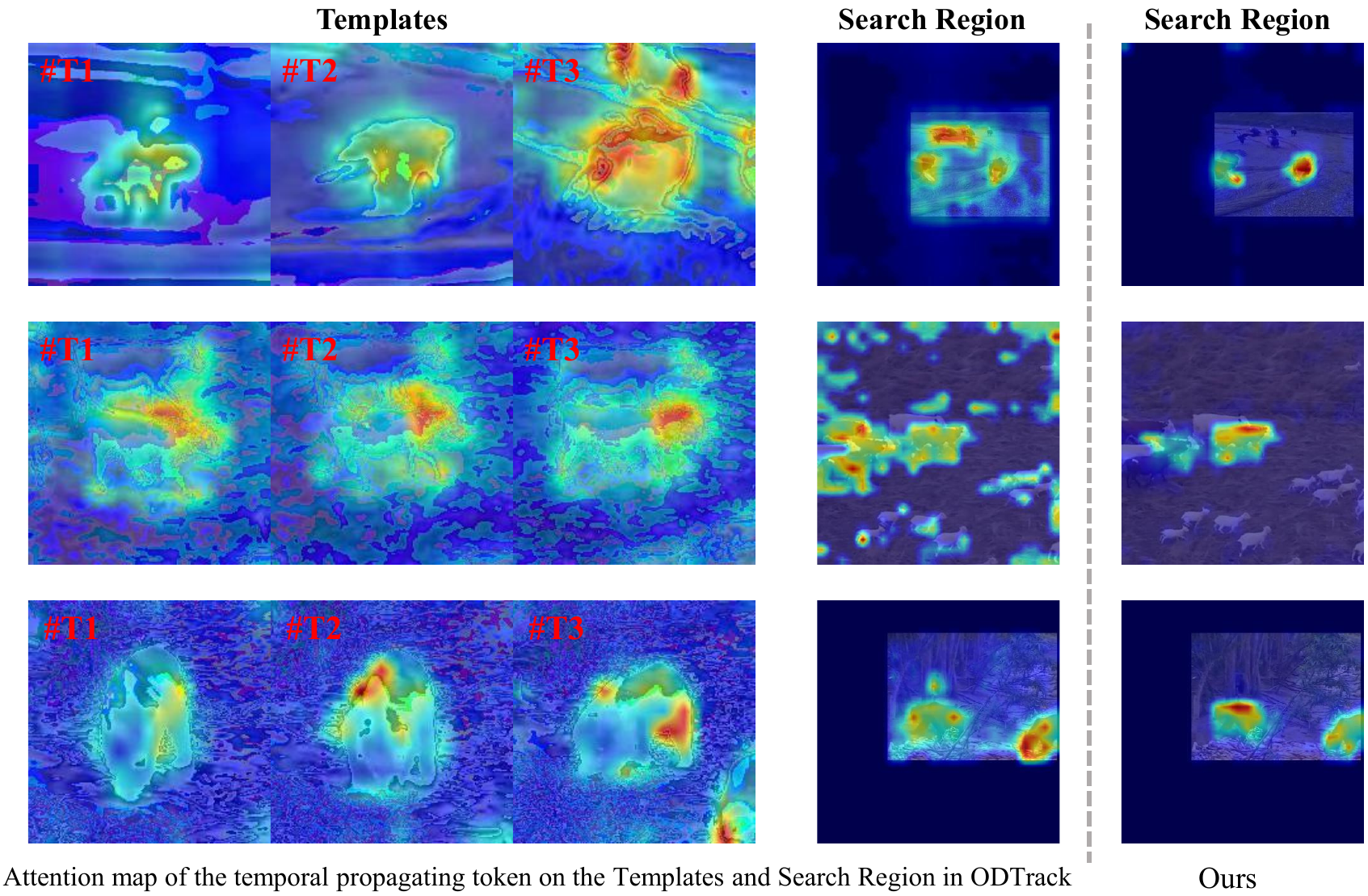}
    \caption{
    Visualization of the attention of the Temporal Propagate Token on the template and search region in ODTrack, compared with the visualization of the attention between the search region and the Dynamic State Feature in DSF module of our method.}
    \label{fig:vis_attn_map}
\end{figure}

\begin{figure}[!t]
    \centering
    \includegraphics[width=0.5\textwidth]{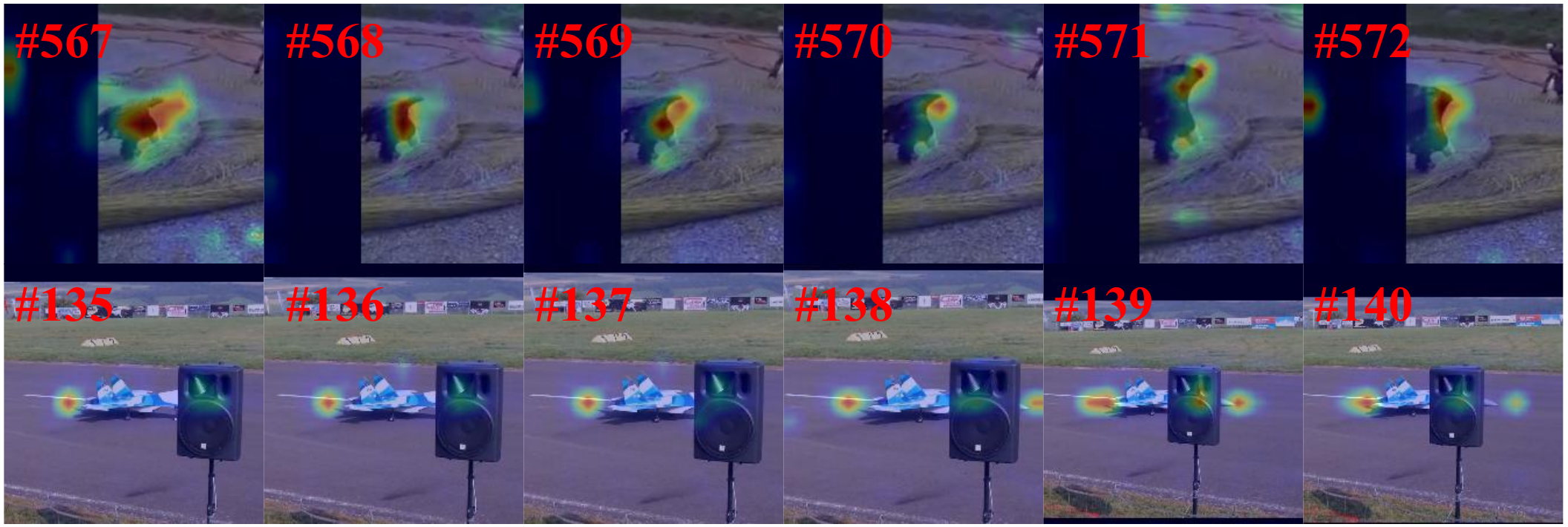}
    \caption{
    Visualization of the attention map between the search region and the Dynamic State Feature output by DSF module when the target undergoes scale variation and fast motion.}
    \label{fig:vis_attn_map_in_sv_and_motion}
\end{figure}

\subsubsection{Results On RGB Visual Tracking} \ 

In order to visually highlight the advantages of our method over existing approaches in challenging scenarios, we provide the detail visualization results in Figure \ref{fig:qualitative}. All videos are from the \emph{test} split of LaSOT. We compare our proposed Uni-MDTrack-B with SUTrack-B\cite{sutrack} and MambaLCT$_{256}$~\cite{li2025mambalct} in terms of performance when the target undergoes sudden movement, deformation, occlusion, and scale variation. All the selected videos are challenging, as described below:

\begin{itemize}
    \item Figure \ref{fig:qualitative}(a) demonstrates the tracking results of three methods when the target suffer from large scale variations.
    \item Figure \ref{fig:qualitative}(b) demonstrates the tracking results of three methods when the targets have partial occlusions and among similar objects.
    \item Figure \ref{fig:qualitative}(c) demonstrates the tracking results of three methods when the target suffers sudden movement or occlusion.
\end{itemize}

We observe that in large scale variations (as shown in Figure \ref{fig:qualitative}(a)), previous trackers struggle to maintain consistent tracking of the correct target. In contrast, our Uni-MDTrack demonstrates superior performance in accurately identifying and consistently tracking the target, even in the presence of sudden movements (as illustrated in Figure \ref{fig:qualitative}(c)). Additionally, in Figure \ref{fig:qualitative}(b), Uni-MDTrack can effectively discriminate the background distractors.

Although our method demonstrates clear advantages over prior approaches, tracking failures can still occur in extremely challenging scenarios. As shown in Figure \ref{fig:vis_bascase}, when the coin is heavily occluded and surrounded by numerous visually similar distractors, all methods eventually fail; however, our method is still able to track the target during the initial stage of occlusion. A similar situation arises in the racing scenario, where the background contains many nearly identical cars that frequently overlap, making the target extremely difficult to predict. In the volleyball scenario, the ball undergoes rapid motion, appears small in scale, and is often blurred, which likewise makes prediction highly challenging. These types of scenes are universally difficult for current tracking models.

\subsubsection{Results on Multi-Modal Visual Tracking} \ 

We further provide visualized comparisons of our proposed Uni-MDTrack against other excellent trackers SUTrack \cite{sutrack}, and STTrack \cite{hu2025exploiting_sttrack} across other modalities including RGB-Depth in Figure \ref{fig:quantitative2}(a), RGB-Event in Figure \ref{fig:quantitative2}(b) and RGB-Thermal in Figure \ref{fig:quantitative2}(c).  Our Uni-MDTrack consistently  exhibit superior performance on these modalities. 

\subsubsection{Attention Map Comparison of Temporal Propagate Token and Dynamic State Feature} \ 


In Figure \ref{fig:vis_attn_map}, we visualized the attention maps for the representative method ODTrack~\cite{ODTrack}, which uses Temporal Propagate Tokens, and compared the attention maps with our approach. For ODTrack, since the Temporal Propagate Token, template, and search region tokens are concatenated together, we can visualize the attention of the Temporal Propagate Token to the template and search region to ascertain what information it actually integrates. On the left side of Figure \ref{fig:vis_attn_map}, we sum and normalize the attention weights of each layer in ODTrack-B, revealing that a significant portion of the Temporal Propagate Token's attention is focused on the templates. This can hinder the token's ability to integrate information about target dynamic state changes. On the right side of Figure \ref{fig:vis_attn_map}, we visualize the cross-attention map of our search region to the Dynamic State Feature in DSF, showing that the attention is more focused and precise.

\subsubsection{Visualization of the dynamic state features output by DSF on the search region when the target undergoes scale variation and fast motion} \ 

To further illustrate how the dynamic target state encoded by DSF evolves when the target undergoes short‑term scale variation or fast motion, we select two representative scenarios in which the target experiences noticeable scale variation and fast movement. As shown in Figure \ref{fig:vis_attn_map_in_sv_and_motion}, we visualize the averaged cross‑attention weights between the outputs of all DSF modules and the search‑region features. Compared with Figure \ref{fig:vis_attn_map}, we increase the transparency of the heatmaps and use consecutive frames to make the target’s motion or appearance changes easier to observe.

From the results in Figure \ref{fig:vis_attn_map_in_sv_and_motion}, for the eagle with unfolding wings, the dynamic features produced by DSF clearly focus on the motion of the wings as they extend. For the fast‑moving airplane, the dynamic features output by DSF clearly focus on the airplane's head-to-tail motion direction. When the airplane's head is occluded, the attention map can roughly estimate the head position, and when the airplane reveals previously occluded parts, DSF can quickly allocate attention to these emerging regions.

\section{Conclusion}

This paper presents a simple, efficient PEFT method for single object tracking, centered on two modules: Memory-Aware Compression Prompt module (MCP) and Dynamic State Fusion module (DSF). These modules achieve a deep fusion of  memory features and continuous dynamic state of the target, enhancing tracking performance while preserving efficiency.
Based on the MCP and DSF modules, we design Uni-MDTrack, which supports five modalities and achieves new state-of-the-art performance as an unified tracker by training 30\% of its parameters. Crucially, both MCP and DSF demonstrate strong generalizability, functioning as effective plug-and-play enhancements for various trackers. We hope this work encourages more low-cost, high-efficiency research in single object tracking.
Meanwhile, we will further explore whether stronger spatio-temporal modeling capabilities can be integrated with reinforcement learning~\cite{huang2026adaptivebatchwisesamplescheduling,huang2026doesreasoningmodelimplicitly,huang2026realtimealignedrewardmodel,zhang2026heterogeneousagentcollaborativereinforcement}, thereby unleashing the spatio-temporal reasoning ability of the tracker.

\bibliographystyle{IEEEtran}
\bibliography{IEEEabrv,reference}

\vfill

\end{document}